\documentclass[final,3p, times]{elsarticle}

\usepackage{amssymb}
\usepackage{amsmath}
\usepackage{url}
\usepackage{color}
\usepackage{algorithm} 
\usepackage{algorithmic} 
\usepackage{amsmath,amssymb,amsfonts} 
\usepackage{mathrsfs} 
\usepackage{graphicx}
\usepackage{textcomp}
\usepackage{relsize} 
\usepackage{cuted}
\usepackage{dblfloatfix} 
\usepackage[bookmarks,bookmarksnumbered]{hyperref} 
\hypersetup{colorlinks = true,
    linkcolor=blue,
    filecolor=magenta,      
    urlcolor=cyan}
\usepackage[labelfont=bf]{caption} 
\usepackage[nameinlink,capitalize]{cleveref} 
\usepackage{flushend} 
\usepackage{booktabs} 
\usepackage[bb=boondox]{mathalfa} 
\usepackage{float} 
\usepackage{makecell} 
\usepackage{setspace}
\usepackage[english]{babel}
\usepackage{framed,multirow}

\definecolor{inchworm}{rgb}{0.7, 0.93, 0.36} 
\definecolor{orng}{rgb}{1.0, 0.31, 0.0} 

\newcommand\eatpunct[1]{} 
\addto\extrasenglish{

} 
\addto\captionsenglish{} 

\journal{Expert Systems with Applications}

\begin{document}







\begin{titlepage}
\doublespacing
{\centering{\Large \textbf{T-MPEDNet: Unveiling the Synergy of Transformer-aware Multiscale Progressive Encoder-Decoder Network with Feature Recalibration for Tumor and Liver Segmentation}}\\
Chandravardhan Singh Raghaw, Jasmer Singh Sanjotra, Mohammad Zia
Ur Rehman, Shubhi Bansal, Shahid Shafi Dar, Nagendra Kumar\\}

\vspace{2em}

{\large Highlights}

\begin{itemize}
\item A novel transformer-aware progressive encoder-decoder for segmenting liver and tumor.
\item Feature recalibration leverages channel-wise dependency to enhance spatial coherence.
\item Multi-scale feature extractor utilize large receptive field for fine-grained feature.
\item Efficiently pinpointing accurate boundaries with morphological erosion operation.
\item Quantitative and qualitative analyses demonstrate the superiority of T-MPEDNet.
\end{itemize}

\vspace{2em}

\noindent This is the preprint version of the accepted paper.\\
\noindent This paper is accepted in \textbf{Biomedical Signal Processing and Control, 2025.}
\\
DOI: \url{https://doi.org/10.1016/j.bspc.2025.108225}
\end{titlepage}


\begin{frontmatter}
\title{T-MPEDNet: Unveiling the Synergy of Transformer-aware Multiscale Progressive Encoder-Decoder Network with Feature Recalibration for Tumor and Liver Segmentation}

\author[1]{Chandravardhan Singh Raghaw}
\ead{phd2201101016@iiti.ac.in}

\author[2]{Jasmer Singh Sanjotra}
\ead{ee220002041@iiti.ac.in}

\author[1]{Mohammad Zia Ur Rehman}
\ead{phd2101201005@iiti.ac.in}

\author[1]{Shubhi Bansal}
\ead{phd2001201007@iiti.ac.in}

\author[1]{Shahid Shafi Dar}
\ead{phd2201201004@iiti.ac.in}

\author[1]{Nagendra Kumar\corref{cor1}}
\ead{nagendra@iiti.ac.in}
\cortext[cor1]{Corresponding author}

\address[1]{Department of Computer Science and Engineering, Indian Institute of Technology (IIT) Indore, Indore 453552, India}

\address[2]{Department of Electrical Engineering, Indian Institute of Technology (IIT) Indore, Indore 453552, India}

\begin{abstract}
Precise and automated segmentation of the liver and its tumor within CT scans plays a pivotal role in swift diagnosis and the development of optimal treatment plans for individuals with liver diseases and malignancies. However, automated liver and tumor segmentation faces significant hurdles arising from the inherent heterogeneity of tumors and the diverse visual characteristics of livers across a broad spectrum of patients. Aiming to address these challenges, we present a novel \textbf{T}ransformer-aware \textbf{M}ultiscale \textbf{P}rogressive \textbf{E}ncoder-\textbf{D}ecoder \textbf{Net}work (T-MPEDNet) for automated segmentation of tumor and liver. T-MPEDNet leverages a deep adaptive features backbone through a progressive encoder-decoder structure, enhanced by skip connections for recalibrating channel-wise features while preserving spatial integrity. A Transformer-inspired dynamic attention mechanism captures long-range contextual relationships within the spatial domain, further enhanced by multi-scale feature utilization for refined local details, leading to accurate prediction. Morphological boundary refinement is then employed to address indistinct boundaries with neighboring organs, capturing finer details and yielding precise boundary labels. The efficacy of T-MPEDNet is comprehensively assessed on two widely utilized public benchmark datasets, LiTS and 3DIRCADb. Extensive quantitative and qualitative analyses demonstrate the superiority of T-MPEDNet compared to twelve state-of-the-art methods. On LiTS, T-MPEDNet achieves outstanding Dice Similarity Coefficients (DSC) of 97.6\% and 89.1\% for liver and tumor segmentation, respectively. Similar performance is observed on 3DIRCADb, with DSCs of 98.3\% and 83.3\% for liver and tumor segmentation, respectively. Our findings prove that T-MPEDNet is an efficacious and reliable framework for automated segmentation of the liver and its tumor in CT scans.

Precise and automated segmentation of the liver and its tumor within CT scans plays a pivotal role in swift diagnosis and the development of optimal treatment plans for individuals with liver diseases and malignancies. However, automated liver and tumor segmentation faces significant hurdles arising from the inherent heterogeneity of tumors and the diverse visual characteristics of livers across a broad spectrum of patients. Aiming to address these challenges, we present a novel Transformer-aware Multiscale Progressive Encoder-Decoder Network (T-MPEDNet) for automated segmentation of tumor and liver. T-MPEDNet leverages a deep adaptive features backbone through a progressive encoder-decoder structure, enhanced by skip connections for recalibrating channel-wise features while preserving spatial integrity. A Transformer-inspired dynamic attention mechanism captures long-range contextual relationships within the spatial domain, further enhanced by multi-scale feature utilization for refined local details, leading to accurate prediction. Morphological boundary refinement is then employed to address indistinct boundaries with neighboring organs, capturing finer details and yielding precise boundary labels. The efficacy of T-MPEDNet is comprehensively assessed on two widely utilized public benchmark datasets, LiTS and 3DIRCADb. Extensive quantitative and qualitative analyses demonstrate the superiority of T-MPEDNet compared to twelve state-of-the-art methods. On LiTS, T-MPEDNet achieves outstanding Dice Similarity Coefficients (DSC) of 97.6
\end{abstract}

\end{frontmatter}

\section{Introduction}
Biomedical image organ segmentation underpins a range of crucial applications, including radiation therapy, computer-aided diagnosis, bio-marker measurement systems, visual augmentation, and surgical navigation~\citep{Ma2021abdomenCT, Valanarasu2022kiunet}. As the largest abdominal organ, the liver holds particular significance in clinical tasks such as surgical planning for diverse liver pathologies, liver volumetry, liver tumor radiotherapy, and living donor liver transplantation~\citep{Li2018hdenseunet}. Computed Tomography (CT) scans continue to be a prominent and widely used method for diagnosing liver and tumors, offering accurate measurements of tumor dimensions, morphology, distribution, and functional hepatic mass, ultimately aiding in precise evaluation of hepatocellular carcinoma and pre-transplant assessment~\citep{Li2018hdenseunet}. Traditionally, radiologists performed slice-by-slice delineation of the liver, and this is a laborious and error-prone approach. Consequently, the development of automated segmentation tools is crucial for the effective diagnosis and treatment of liver and tumor conditions, as they can significantly improve diagnostic workflows.

\subsection{Challenges in Automated Segmentation of the Liver and Tumors}
\begin{figure*}[!ht]
  \centering
  \includegraphics[width=\textwidth]{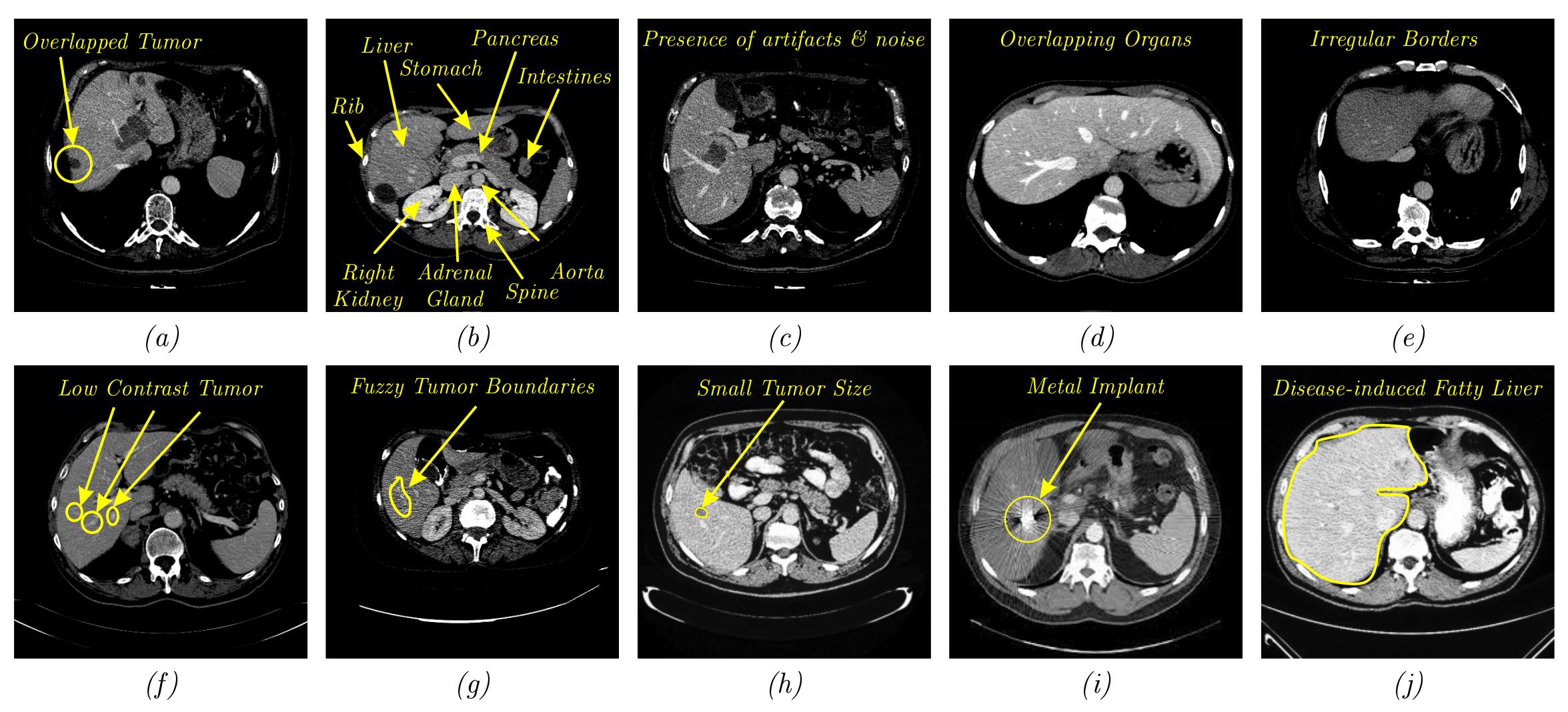}
  \caption{Diverse CT scans labeled (a) to (j) highlighting liver segmentation complexities, including (a) overlapping organs and tumors, (b) intricate liver anatomy, (c) the presence of artifacts and noise, (d) overlapping intensities, (e) irregular borders due to cirrhosis, (f) low contrast between the liver and surrounding healthy tissues, (g) fuzzy and indistinct tumor boundaries, (h) limited visibility of small tumors, (i) streaking artifacts caused by metal implants, and (j) alterations in liver intensity and texture due to disease.}
  \label{fig:intro}
\end{figure*}

Accurately segmenting the liver in contrast-enhanced CT scans remains a significant challenge arising from faint intensity differences between surrounding organs and the liver. Radiologists often employ injection protocols to enhance CT scans for tumor visualization, potentially introducing noise in the liver region. Compared to liver segmentation, automatic tumor segmentation presents even greater difficulties arising from several factors. One challenge lies in the tumor's inherent variability – dimensions, morphology, location, and multiplicity – within individual patients, which confounds automated segmentation. Additionally, tumors present indistinct boundaries with neighboring organs, further limiting the effectiveness of segmentation algorithms. Furthermore, the inherent anisotropy of many CT scans, manifested by significant voxel spacing variations along the z-axis, introduces additional complexities for automated segmentation. Finally, scanner variations and CT phases induce substantial discrepancies in organ appearances, as exemplified in~\autoref{fig:intro}.

\subsection{Exploring the Research Trajectory}
To tackle the above challenges, previous attempts at liver tumor segmentation through conventional methods, including region growing~\citep{Yang2021region}, deformable models~\citep{He2024deform}, threshold-based methods~\citep{Liu2022threshold} and, fuzzy C-means clustering~\citep{Dickson2024sparse}, while reducing human dependence, remain restricted by handcrafted features. This limitation hinders their ability to capture complex representations and leads to substandard segmentation results~\citep{Di2023tdnet}. The transformative impact of deep learning on healthcare has led to the development of numerous liver tumor segmentation techniques based on Convolutional Neural Networks (CNNs). A pivotal advancement in bio-medical image segmentation was realized with U-Net~\citep{Ronneberger2015unet}, which marked a paradigm shift by introducing skip connections, revolutionizing training and feature quality for segmentation tasks. Building upon this breakthrough, CNN-based techniques have become the driving force in modern bio-medical image segmentation.

Building upon the success of U-Net, researchers have proposed numerous diverse networks, further advancing liver tumor segmentation. Liao et al.~\cite{Liao2024mscfunet} introduce a lightweight multi-scale context fusion strategy incorporating dual attention to enhance segmentation performance, while Tu et al.~\cite{Tu2023slice} introduced a novel feature-fusing method incorporating adjacent tissues to reduce false positives. Li et al.~\cite{Li2023ace} employed an attention-based context encoding module to extract high-fidelity features for precise liver surface segmentation. Kushnure et al.~\cite{Kushnure2022hfru} modified the skip pathways of U-Net, integrating atrous spatial pyramid pooling with the squeeze-and-excitation network to achieve effective segmentation of livers and tumors in CT scans. Jiang et al.~\cite{Jiang2023rmau} addressed gradient disappearance through a tailored model for feature recalibration and handling interdependencies, boosting segmentation performance. More recently, the attention-based architecture of Transformers~\citep{Liu2022transunetp, Xie2021cotr, Vaswani2017} has garnered significant interest in both natural and medical image tasks, demonstrating remarkable performance. However, their application in liver tumor segmentation remains relatively unexplored.

\subsection{Gaps in the Literature}
Small structure segmentation remains a challenge for deep convolutional networks due to inherent limitations. As network depth increases, convolutions prioritize extracting high-level spatial features, neglecting the crucial relationship between local and global features of small structures. This solely convolutional approach hampers the network's understanding of the overall structure. Additionally, traditional concatenation or aggregation approaches need to capture the diverse nature of features and limit the receptive field, ultimately degrading segmentation performance. Furthermore, high-resolution edge information, vital for accurate segmentation, is often lost during feature extraction.

While U-Net and its variants~\citep{Zhang2024fafsunet} excel at segmenting large structures, they falter when faced with minor or noisy objects, as demonstrated in~\autoref{fig:intro}. Despite skip connections facilitating local information flow to the decoder, our experiments revealed their inefficiency in accurately segmenting small anatomical landmarks with blurred boundaries.

Integrating transformers could offer a potential solution for acquiring global features. However, the discrete and irregular nature of liver and tumors presents a significant challenge for transformer-based approaches, hindering their ability to locate and extract global features accurately. Therefore, developing a novel approach that effectively unlocks the potential of spatial-global feature learning could revolutionize diagnosing and treating liver diseases by enabling accurate segmentation of complex liver and tumors.

\subsection{Contributions of the Proposed Work}
We tackle the limitations of previous work in handling long-range dependencies and multi-scale features by presenting a novel architecture, Transformer-aware Multiscale Progressive Encoder-Decoder Network (T-MPEDNet), for automated liver and tumor segmentation. T-MPEDNet leverages a synergistic architecture featuring a progressive encoder-decoder utilizing skip connections for retaining spatial information and a transformer-based dynamic attention mechanism for capturing global patterns. This enables T-MPEDNet to extract fine-grained high-level information via convolutions while simultaneously learning global context through Transformer attention. Furthermore, we incorporate a boundary refinement module to enhance the accuracy of liver and tumor boundary recognition. The efficacy of T-MPEDNet is demonstrated by achieving state-of-the-art results on two publicly available datasets: Liver Tumor Segmentation Challenge (LiTS) dataset~\citep{Bilic2023lits} and 3D Image Reconstruction for Comparison of Algorithm Database (3DIRCADb) dataset~\citep{soler20103d}. Our key contributions are summarized as follows:

\begin{itemize}
    \item We propose a novel framework, the Transformer-aware Multiscale Progressive Encoder-Decoder Network (T-MPEDNet), for segmenting liver and tumors. T-MPEDNet leverages the power of a Progressive Encoder-Decoder architecture composed of Adaptive Feature Extractor (AdaFEx) modules and a Compressive Channel Recalibration (CCR) module. The architecture adaptively recalibrates fused channel-wise features via skip connections to learn fine-grained details and enhance representation capability while preserving spatial integrity.
    \item We introduce a transformer-inspired Dynamic Contextual Attention (DCA) module after the encoder block. DCA module unifies dynamic convolutions before multi-headed attention to efficiently capture deep global features and learn contextual relationships among spatial information.
    \item We integrate a novel Multi-Scale Atrous Spatial (MSAS) stacked module, utilizing multiple levels of atrous convolution. This module effectively learns local detail features and ensures a comprehensive collection of global features and contextual information, leading to more accurate and reliable predictions.
    \item We propose a Morphological Boundary Refinement (MBR) module that utilizes a binary mask and corresponding hierarchical features to enrich boundary detail information. This significantly improves segmentation accuracy by sharpening edges and delineating boundaries precisely.
    \item T-MPEDNet surpasses twelve state-of-the-art liver and tumor segmentation methods on two public datasets. The model demonstrates an edge over existing approaches in segmenting tiny tumors and complex hepatic structures, highlighting its potential for clinical applications.
\end{itemize}

\subsection{Outline of the Paper and Approach}
This article thoroughly examines the complexities of automated liver and tumor segmentation from CT scans, focusing on two key challenges: (a) handling the diverse and often subtle visual characteristics of lesions and tumors across a wide range of patients and (b) improving segmentation accuracy, especially for small and poorly defined lesions. To address these issues, we introduce a novel Transformer-aware Multiscale Progressive Encoder-Decoder Network (T-MPEDNet). \autoref{sec:related} comprehensively explore existing literature on liver and tumor segmentation methods. In~\autoref{sec:methodology}, we introduce T-MPEDNet, a novel framework that leverages a progressive encoder-decoder architecture to extract deep fine-grained features from CT scans. T-MPEDNet's core lies in its Progressive encoder-decoder architecture (refer~\autoref{sec:proende}). This architecture progressively extracts deep context from low-level details, transitioning to higher-level semantics. The Adaptive Feature Extraction (AdaFEx) module (\hyperlink{adafex}{Section 3.2.2.1}) adaptively extracts deep features, while the Compressive Channel Recalibration (CCR) module (\hyperlink{ccr}{Section 3.2.2.2}) refines their importance. The Dynamic Contextual Attention (DCA) module (\hyperlink{dca}{Section 3.2.2.3}) captures global features while preserving spatial integrity. Finally, the Multi-Scale Atrous Spatial (MSAS) module (\hyperlink{msas}{Section 3.2.2.4}) enhances low-level feature representation with intricate details using varying receptive fields. \autoref{sec:boundryrefine} introduces the Morphological Boundary Refinement (MBR) module, which enriches boundary detail information and improves segmentation accuracy by precisely sharpening edges and delineating boundaries. \autoref{sec:experiment} outlines the rigorous evaluation process conducted during the research. \autoref{sec:discussion} discusses the importance of the proposed framework and outlines potential future research directions. Finally,~\autoref{sec:conclusion} summarizes the essential findings and explores potential avenues for future research.

\section{Related Work}
\label{sec:related}
This section aims to deliver an exhaustive literature review of prior research analysis conducted in the domain of liver and tumor segmentation methodologies. To facilitate the analysis, existing research is broadly classified into three categories: Encoder-Decoder-based segmentation, Attention-based segmentation, and Multi-Scale Feature Extractor-based segmentation.

\subsection{Encoder Decoder-based Segmentation}
\label{sec:rwcnn}
Deep learning architectures such as U-Net~\citep{Ronneberger2015unet}, built upon the encoder-decoder paradigm, have revolutionized the field of biomedical image segmentation~\citep{Bilic2023lits, Gopinath2023, Mahendran2023, Soomro2023, Qayyum2024coattunet}. Di et al.~\cite{Di2022autolt} presented a novel liver tumor extraction framework from Computed Tomography scans using 3D U-Net. The 3D U-Net was used for liver segmentation, followed by a hierarchical iterative superpixel approach to refine tumor extraction. Chen et al.~\cite{Chen2023draunet} unveiled a method that used multiple planes for liver segmentation, involving coronal Computed Tomography (CT) slices along with transverse slices, thus incorporating more 3D spatial information. Tu et al.~\cite{Tu2023slice} devised a slice-fusion method that incorporated the long-range anatomical coherence between tissues and the CT scans, analyzed the features of neighboring slides, and fused them to form the segmentation map.

Chen et al.~\cite{ChenKiu2023} proposed an improved variant of 3D KiU-Net~\citep{Valanarasu2022kiunet}, which incorporated a Kite-Net branch powered by a transformer~\citep{Vaswani2017}, resulting in an over-complete architecture that resulted in more detailed features for small structures. LI et al.~\cite{LI2020} constructed a U-Net variant with a novel feature supervision approach that employs a low-dimensional representation of extracted features for improved segmentation. Although powerful, the above methodologies are unable to capture boundary-level details as the liver presents intricate pathological features. To accommodate this substantial variation, we proposed a custom encoder-decoder architecture, constituting multiple convolutional layers with varying kernel sizes, enhancing global detail acquisition.

\subsection{Attention-based Segmentation}
\label{sec:rwhybrid}
Medical image segmentation has seen a surge in the use of attention-based architectures for improved performance. Non-local self-attention models have mainly performed well due to their prowess in capturing long-range dependencies, crucial for the precise delineation of intricate tumor boundaries~\citep{Zhao2022, Li2023sdmt, Tomar2023, Wu2023}. Li et al.~\cite{Li2023ace} introduced an innovative framework termed the Attentive Context-Enhanced Network (AC-E Network), comprising an Attentive Context Encoding Module (ACEM). The ACEM incorporates 3D context with minimal parameter expansion, while the dual segmentation branch with complementary losses optimizes region and boundary localization.

Kushnure et al.~\cite{Kushnure2022hfru} proposed the High-Level Feature Fusion and Recalibration UNet (HFRU-Net), which entails the modification of U-Net skip pathways through the incorporation of local feature reconstruction and a feature fusion mechanism. HFRU-Net also deployed Squeeze and Excitation Network (SENet) modules, which dynamically adjust channel weights, amplifying informative features and suppressing less relevant ones. Di et al.~\cite{Di2023tdnet} outlined TD-Net, comprising of a direction guidance block, skip connections, along with dual decoders and a shared encoder. Leveraging the shared encoder for efficient feature extraction, TD-Net utilized dual decoding branches: generating an initial segmentation map and predicting direction information for iterative refinement. Li et al.~\cite{Li2022litsnet} devised a framework named LITSNet, incorporating the Shift-Channel Attention Module. This Module is designed to model the inter-dependency amongst the features in adjacent channels for better performance. In addition, the Weighted-Region (WR) loss function was also utilized. Liu et al.~\cite{Liu2023} presented a Global Context and Hybrid Attention Network (GCHA-Net) that leveraged a Global Attention Module (GAM) and a Feature Aggregation Module (FAM), incorporating a Local Attention Module (LAM) to capture the necessary spatial information. Alam et al.~\cite{Alam2023} employed dilated convolution blocks within the U-Net architecture to capture multi-scale context features characterized by diverse receptive field sizes. Additionally, they integrated a dilated inception block between the encoder and decoder paths to mitigate issues related to information recession and address the semantic gap between features. Compared to the above existing techniques, our proposed T-MPEDNet harnesses the strengths of a transformer-inspired attention module. This module facilitates the integration of long-range contextual information with dynamically extracted spatial features, enabling the identification of intricate hepatic characteristics with superior accuracy.

\subsection{Multi-Scale Feature Extractor-based Segmentation}
\label{sec:rwtransformer}
This section discusses methods that exploit the intrinsic multi-scale characteristics of medical images by analyzing images at different scaling levels. This technique has been used across a variety of domains, demonstrating remarkable results~\citep{Cui2021, Du2021, Abdar2023, Yue2023, Dar2025}. Jianget et al.~\cite{Jiang2023rmau} proposed a novel Residual Multi-scale Attention U-Net (RMAU-Net) utilizing inter-spatial information along with residual connections for liver and tumor segmentation from CT scans. Lei et al.~\cite{Lei2022defednet} outlined the Deformable Encoder-Decoder network (DefED-Net) featuring deformable convolution and a novel Ladder-Atrous-Spatial-Pyramid-Pooling (Ladder-ASPP) module for enhanced feature representation and ultimately better segmentation. Chen et al.~\cite{Chen2022fraunet} presented an automated dual-step liver and tumor segmentation method, incorporating a Fractal U-Net (FRA-Unet), which utilized a cascading framework and a fully connected Conditional Random Field (CRF) to segment and refine the final output. 

Valanarasu et al.~\cite{Valanarasu2022kiunet} suggested a convolutional architecture in which the input image was projected to a higher dimension, and the receptive field was constrained from increasing in the deep layers. Chen et al.~\cite{Chen2023msfanet} introduced Multi-Scale Atrous Downsampling in addition to Residual Attention block in the encoder of Multi-scale Feature Attention Network to learn diverse tumor features and extract them at variating scales simultaneously. Lv et al.~\cite{Lv2022} employed a dual attention gate in each skip connection and an atrous encoder within their Cascaded Atrous Dual-Attention U-Net segmentation approach, allowing it to extract more comprehensive contextual features from CT scans compared to the standard encoder. The above methodologies have shown remarkable results, yet they must improve in certain areas. On the other hand, T-MPEDNet leverages the parallel atrous convolution filters with broader receptive field size to capture fine-grained multi-scale features, thereby enhancing the efficiency of feature extraction.

\section{Methodology}
\label{sec:methodology}
The initial section of this paper, \autoref{sec:formulationoverview}, comprehensively outlines the problem formulation and provides a glimpse of the proposed T-MPEDNet. \autoref{sec:tmpednetmethod} then undertakes a thorough examination of T-MPEDNet, meticulously illustrating its internal components. The reformation block is the first to be introduced in detail in Section~\autoref{sec:reform}, followed by an in-depth analysis of the progressive feature encoder-decoder in~\autoref{sec:proende}. Finally,~\autoref{sec:boundryrefine} concludes the detailed exposition of T-MPEDNet by elucidating the morphological boundary refinement module.

\begin{figure*}[!b]
  \centering
  \includegraphics[width=\textwidth]{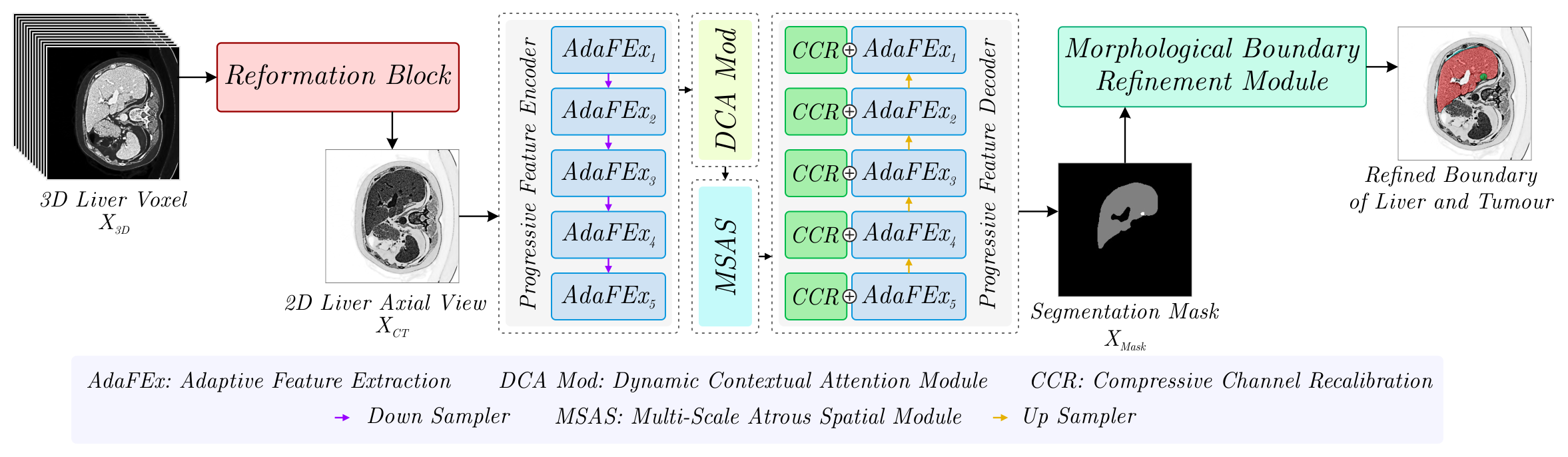}
  \caption{The figure outlines the proposed framework for precisely segmenting liver and tumors. The process begins with a 3D liver voxel input, which transforms 2D slices, denoted as $X_{CT}$, through the reformation block, resulting in enhanced overall quality. A multi-stage encoder extracts progressively deeper features, guided by Adaptive Feature Extraction (AdaFEx) modules. The Dynamic Contextual Attention (DCA) module synergizes with the Multi-Scale Atrous Spatial Pooling (MSAS) module to acquire global multi-scale features. The progressive feature decoder integrates with Channel-Style Recalibration (CSR) to realign displaced information from lower-level stages and generate precise segmentation masks. Finally, morphological boundary refinement refines the final boundary mask to adjust the indistinct boundary of the liver and tumor.}
  \label{fig:tmpednet_floorplan}
\end{figure*}

\subsection{Problem Formulation and Method Overview}
\label{sec:formulationoverview}
We aim to segment the liver and tumor in the CT scan image by precisely predicting the pixel-level mask of the liver and any tumors present in a CT scan image. The goal is to process the input $X_{CT} \in \mathbb{R}^{H\times W \times C}$ drawn from the labeled source dataset $\mathfrak{D}_s$ which has a spatial size of $H \times W$ and $C$ channels. The objective is to generate a corresponding 2D binary mask $Y\in \{0,1,2\}^{H\times W \times C}$ where labels are assigned to each pixel: $0$ for the background, $1$ for the liver, and $2$ for a tumor. Here, $\mathfrak{D}_s$ represents the source labeled dataset that can generalize effectively to an unseen target domain $\mathfrak{D}_t$. The segmentation task is accomplished by training the segmentation model $\mathcal{S}_{\theta}:X_{CT} \rightarrow Y$ on source dataset $\mathfrak{D}_s$. Our approach in this study is to minimize the dice loss $\mathcal {L}$ with $\theta$ as a model parameter, which can be formally expressed in~\cref{eq:diceloss}:

\begin{equation}
\label{eq:diceloss}
\min _{{\theta }} \mathbb {E}_{(X_{CT},Y) \sim \mathfrak{D}_s} \Bigl[ \mathcal {L} \bigl(\mathcal{S}(X_{CT}; \theta), Y \bigr)\Bigr] 
\end{equation}

\autoref{fig:tmpednet_floorplan} illustrates the architectural blueprint of the T-MPEDNet framework, which aims to facilitate the segmentation performance for both liver and tumor regions. The T-MPEDNet framework is equipped with three core components: (a) a reformation block that assists the progressive feature encoder-decoder structure by eliminating irrelevant details from the CT volume and enhancing data quality through normalization techniques to optimize the network learning rate, facilitating the capture of deep features; (b) a progressive feature encoder-decoder structure designed to extract deep feature map, incorporating dynamic contextual attention $\delta_{DCA}$ with a multi-scale atrous spatial module $\mu_{MSAS}$. $\delta_{DCA}$ globally analyzes the deep features from a semantic viewpoint to underscore effective channels, while $\mu_{MSAS}$ amplifies valuable features across both global and local scales. Finally, (c) the morphological boundary refinement $\beta_{MBR}$ sharpens ambiguous boundaries by employing morphological erosion operations. These operations refine blurred edges and grid shapes, resulting in dense and smooth liver and tumor boundaries. Next,~\autoref{sec:tmpednetmethod} focuses on a thorough explanation of the main components of our framework.

\subsection{Transformer-aware Multiscale Progressive Encoder-Decoder Network}
\label{sec:tmpednetmethod}

\subsubsection{Reformation Block}
\label{sec:reform}
\begin{figure}[!ht]
  \centering
  \includegraphics[width=0.5\textwidth]{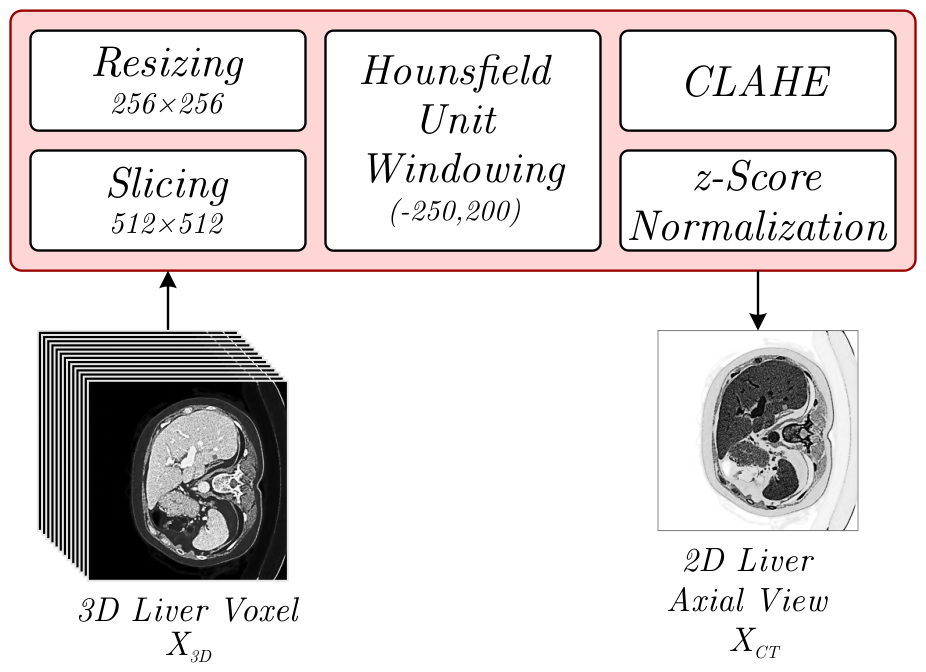}
  \caption{Visualization of the processing pipeline within the Reformation Block.}
  \label{fig:tmpednet_refb}
\end{figure}

The reformation block systematically employs a sequence of processing steps to remove unwanted elements from the CT volume, thereby facilitating liver area segmentation, as outlined in~\autoref{fig:tmpednet_refb}. We initiated the process by converting the 3D liver voxel $X_{3D}$ into 2D slices of $512 \times 512$ pixels, resizing to $256 \times 256$ CT scans to optimize computational efficiency. Hounsfield Units (HU) windowing, with a window size of $(-250, 200)$, excludes irrelevant details from the CT images. HU, a standardized unit for quantifying the relative densities of internal body organs, typically assigns liver tissue radiodensities ranging from $40$ to $50$ HU. Following this, we applied Contrast Limited Adaptive Histogram Equalization (CLAHE) to prevent noise over-amplification, enhance local contrast, and improve organ detail visibility. Finally, the z-score normalization technique is applied to standardize pixel intensities and minimize variations arising from acquisition differences.

\subsubsection{Progressive Feature Encoder Decoder}
\label{sec:proende}
\begin{figure*}[!ht]
  \centering
  \includegraphics[width=\textwidth]{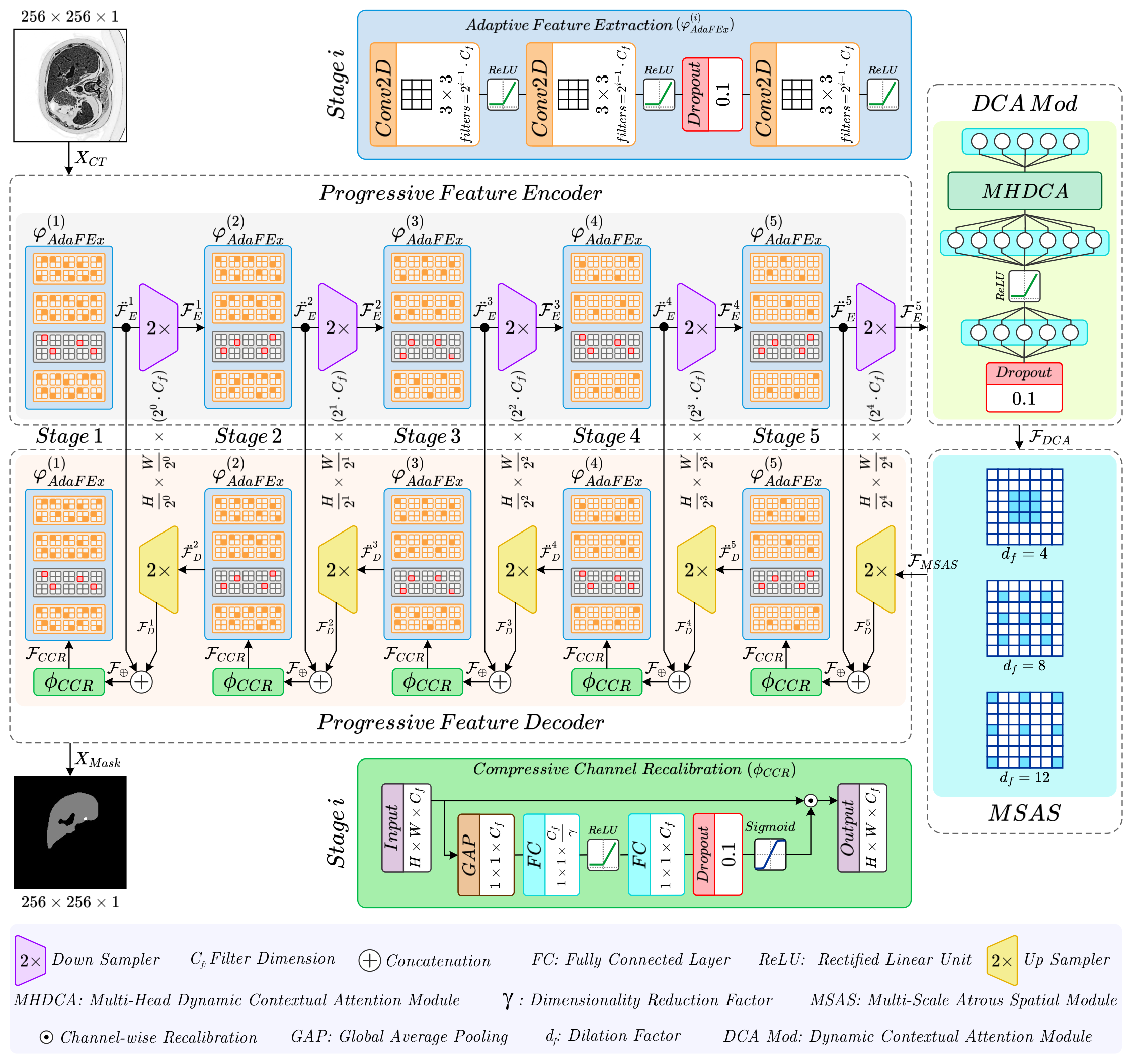}
  \caption{Architecture of the Progressive Feature Encoder-Decoder (PED) block, a core component of the proposed T-MPEDNet framework. PED leverages five mirrored Adaptive Feature Extraction (AdaFEx) modules, a Compressive Channel Recalibration (CCR) module, a Dynamic Contextual Attention (DCA) module, and a Multi-Scale Atrous Spatial (MSAS) module to progressively extract and refine features for accurate liver and tumor segmentation. The block receives a CT scan $X_{CT}$ with a spatial dimension of $(256 \times 256 \times 1)$ as input and generates a binary segmentation mask $X_{Mask}$ of the same dimensions.}
  \label{fig:tmpednet_proposed}
\end{figure*}

\autoref{fig:tmpednet_proposed} outlines the architecture of the progressive feature encoder-decoder block integrated into the proposed T-MPEDNet for liver and tumor segmentation in CT scans. The proposed encoder-decoder module aims to (a) efficiently extract features from input CT scans, (b) adaptively recalibrate fused features on a channel-wise basis, (c) preserve spatial integrity to extract higher-level global features, and (d) capture diverse multi-scale features with varying receptive fields to enhance low-level feature representation with intricate spatial details. To achieve these objectives, the encoder-decoder network integrates four key components: the Adaptive Feature Extraction (AdaFEx) module,  the Compressive Channel Recalibration (CCR) module used in the decoder network, the Dynamic Contextual Attention module, and the Multi-Scale Atrous Spatial (MSAS) module.

The progressive encoder is composed of five AdaFEx modules, labeled as $\varphi_{AdaFEx}^{(i)},\,i\in[1,2,\ldots,5]$, as illustrated in~\autoref{fig:tmpednet_proposed}. The $\varphi_{AdaFEx}$ takes a liver CT scan $X_{CT}$ with dimensions $H \times W \times 1$ as input, where $H$ is the height, $W$ is the width, and $1$ is the grayscale channel. Adaptive deep features, denoted as $\mathcal{\ddot{F}}_E^i$, are extracted for each AdaFEx module, followed by a down-sampling operation. The down-sampler, denoted as $\bigvee$, incorporates a max pooling layer with a $2 \times 2$ pooling window, reducing the spatial dimensions by half while preserving the most salient features. The output after down-sampling are labeled as $\mathcal{F}_E^1,\mathcal{F}_E^2,\ldots,\mathcal{F}_E^5$. The progressive encoder block computes the final feature map by employing a down-sampler, as formally expressed in~\cref{eq:proencoder}:

\begin{equation}
\label{eq:proencoder}
\mathcal{F}_E^i= \bigvee \left(\varphi_{AdaFEx}^{(i)}(\mathcal{F}_E^{i-1};\theta_E^i)\right),
\quad i\in \{1,2,\ldots,5\}
\end{equation}

Here, $\theta_E^i$ denotes the parameters of the $i$-th AdaFEx module $\varphi_{AdaFEx}^{(i)}$, $\mathcal{F}_E^{i}$ is the feature map generated by the down-sampler integrated with $\varphi_{AdaFEx}^{(i)}$, and $\mathcal{F}_E^{0}=X_{CT}$ is the input image.

The progressive encoder generates spatially rich feature maps, feeding them into the Dynamic Contextual Attention (DCA) module. The DCA actively learns a dynamic attention matrix, leveraging the embedded context and amplifying visual representation capabilities. The resulting fusion of static and dynamic contextual information is denoted as $\mathcal{F}_{DCA}$ and is fed into the Multi-Scale Atrous Spatial (MSAS) module. This module actively extracts multi-scale features with diverse receptive fields, effectively capturing the rich spatial details within low-level features. The resulting multi-scaled output, represented as $\mathcal{F}_{MSAS}$, is the input to the progressive decoder.

The progressive feature decoder refines the multi-scale feature map (denoted as $\mathcal{F}_{MSAS}$) by upsampling it and generating segmentation masks. The progressive feature decoder utilizes encoder features to guide the upsampling process through five mirrored AdaFEx modules, each replicating the encoder's block architecture.

As input, each AdaFEx module receives recalibrated features, denoted as $\phi_{CCR}^{(i)} \in [1,2,\ldots,5]$ from the Compressive Channel Recalibration (CCR) module. The CCR module combines up-sampled reconstructed features (denoted as $\mathcal{F}_D^{i}$) from the previous $\varphi_{AdaFEx}^{(i+1)}$ module with the output of the corresponding AdaFEx module from the mirrored encoder. A deconvolution layer with $2 \times 2$ kernel size acts as the up-sampler ($\bigwedge$), enhancing the dimensionality of the feature maps across spatial dimensions and enabling the network to generate high-resolution segmentation masks. Each AdaFEx module in the decoder outputs a refined feature map, denoted as $\mathcal{\ddot{F}}_D^1,\mathcal{\ddot{F}}_D^2,\ldots,\mathcal{\ddot{F}}_D^5$. The final feature map is generated by the last AdaFEx module, and its computation is formally expressed in~\cref{eq:prodecoder}. Finally, the progressive decoder outputs a 2D binary mask of the liver and the corresponding tumor. 

\begin{equation}
\label{eq:prodecoder}
\mathcal{\ddot{F}}_D^i= \varphi_{AdaFEx}^{(i)}\Bigl(\phi_{CCR}^{(i)}\bigl(\mathcal{\ddot{F}}_E^i \, \oplus \,\bigwedge(\mathcal{\ddot{F}}_D^{i+1})\bigr);\theta_D^i\Bigr)\, i\in \{1,2,\ldots,5\}
\end{equation}

Here, $\oplus$ representing the concatenation, $\theta_D^i$ represents the parameters of the $i$-th AdaFEx module $\varphi_{AdaFEx}^{(i)}$, and  $\mathcal{\ddot{F}}_D^i$ is the feature map produced by the AdaFEx integrated with $\phi_{CCR}^{(i)}$. Also, $\mathcal{\ddot{F}}_D^5=\mathcal{F}_E^5$ and $\mathcal{\ddot{F}}_D^1=X_{Mask}$ is the final 2D-mask of the liver and corresponding tumor.

The progressive encoder-decoder architecture employs skip connections between corresponding AdaFEx modules. These connections bridge the gap between feature maps with decreased spatial resolution and increased spatial resolution, alleviating information loss during downsampling in the encoder. During upsampling in the decoder, the skip connections provide a rich source of contextual information for the higher-resolution layers, empowering them to generate finer details in the final segmentation mask.

\hypertarget{adafex}{\paragraph[]{(i) Adaptive Feature Extraction Module:\eatpunct}}

Neural networks leverage the feature extraction capabilities of convolution layers, representing a powerful approach for learning data characteristics. However, medical image analysis remains a significant challenge despite advancements in the field, primarily due to the diverse types of medical images demanding different feature extraction strategies. This work introduces Adaptive Feature Extraction (AdaFEx), a novel feature extraction module denoted as $\varphi_{AdaFEx}$. AdaFEx efficiently learns salient features of liver and tumor tissues from CT scans, empowering the model to locate and segment these structures with high precision.

AdaFEx progressively extracts deep context from low-level details, gradually transitioning to higher-level semantic representations. As illustrated in~\autoref{fig:tmpednet_proposed}, AdaFEx takes the CT scan image $(X_{CT})$ as input and feeds it through five stages $\left.\varphi_{AdaFEx}^{(i)}\right|_{i=1}^5$ takes $X_{CT}$. Each stage comprises three Conv layers with $f$ filters of $3 \times 3$ kernel, where $f$ progressively increases at each stage according to the formula $2^{i-1} \cdot C_f$, where the value of $C_f$ is $16$. ReLU activation functions following each Conv layer introduce non-linearity, enabling the model to extract and learn hidden complex connections and relationships within the data. A Dropout layer is integrated after the second Conv layer in each stage to enhance regularization and reduce neuron interdependencies. 

\begin{figure}[!ht]
    \centering    
    \includegraphics[width=0.5\textwidth]{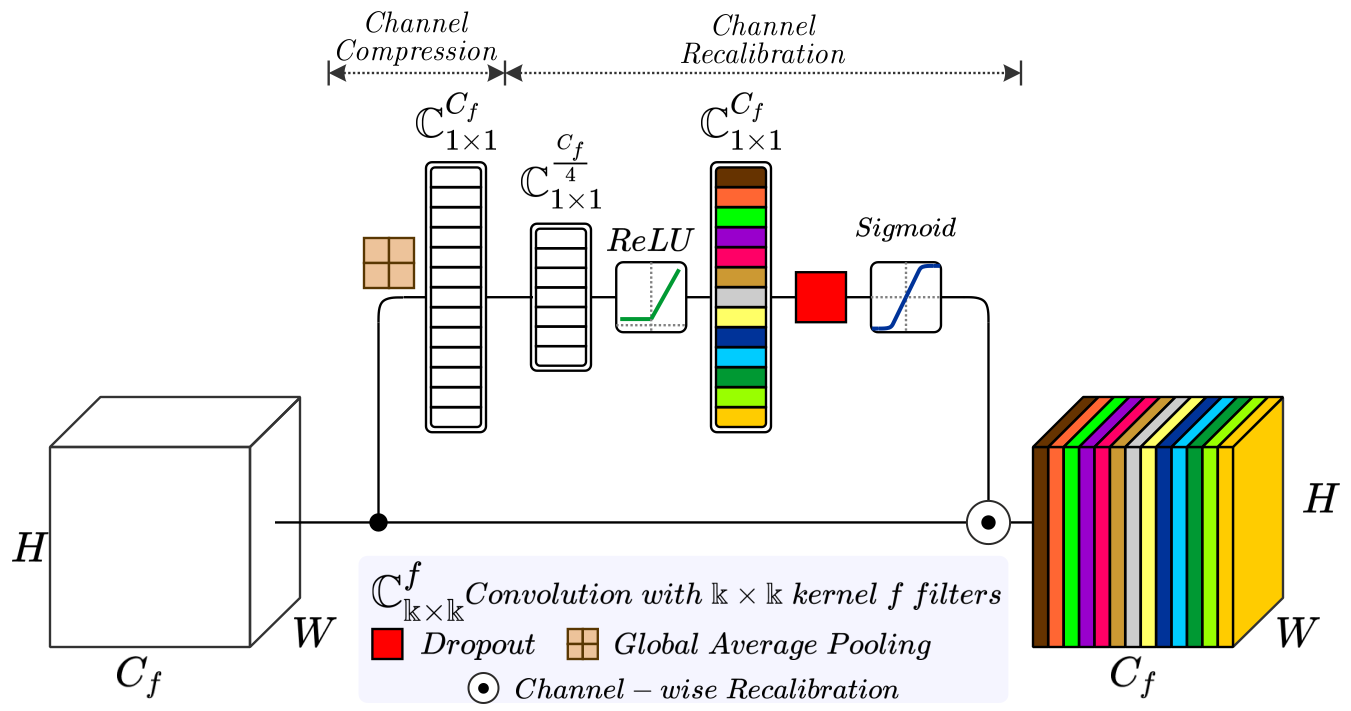}
    \caption {The Architecture of the Compressive Channel Recalibration module.}
    \label{fig:tmpednet_ccr}
\end{figure}

\hypertarget{ccr}{\paragraph[]{(ii) Compressive Channel Recalibration Module:\eatpunct}}
The Compressive Channel Recalibration (CCR) module, denoted as $\phi_{CCR}$, effectively leverages channel-wise dependencies in feature maps. It automatically acquires weights and recalibrates original feature channels, enhancing feature representation. The detailed internal structure of the CCR module is visualized in~\autoref{fig:tmpednet_ccr}. It first compresses concatenated feature map $\mathcal{F}_{\oplus}\in \mathbb{R}^{H \times W \times C_f}$ through channel aggregation, producing a compressed channel representation. Subsequently, a recalibration operation generates modulation weights for channels, which are applied to $\mathcal{F}_{\oplus}$ to yield the output $\mathcal{F}_{CCR}$. This output then serves as input to the AdaFEx module of the progressive decoder block.

For an input $\mathcal{F}_{\oplus}=[f_{\oplus}^1, f_{\oplus}^2,\ldots,f_{\oplus}^{C_f}]$, where $f_{\oplus}^k\in \mathbb{R}^{H \times W \times C_f}$, we leverage global average pooling to reduce the spatial dimensions of each channel while preserving channel-specific information. This produces a static vector $\mathcal{Z} \in \mathbb{R}^{1 \times 1 \times C_f}$ by by shrinking $\mathcal{F}_{\oplus}$ along its spatial dimensions $H \times W$. The \textit{k}-th element of $\mathcal{Z}$ is calculated by:

\begin{equation}
\label{eq:ccrcompression}
z_k=\mathcal{F}_{com}(f_{\oplus}^k)=\sum_{i=1}^{H \times C} f_{\oplus}^k(j)
\end{equation}

The above procedure embeds global spatial information within a vector $z_k\in\mathcal{Z}$. Following the compression operation, the recalibration operation ($\mathcal{F}_{rec}$) is calculated to capture channel-wise dependencies comprehensively. Recalibration involves transforming $\mathcal{Z}$ to $\hat{\mathcal{Z}}$ by integrating two Fully Connected (FC) layers as outlined in~\cref{eq:ccrrecalibration}: 

\begin{equation}
\label{eq:ccrrecalibration}
\mathcal{A}=\mathcal{F}_{rec} (\mathcal{Z}, \mathcal{W})=\underset{Sigmoid}{\sigma}\Bigl( \bigl(\mathcal{W}_2 \underset{ReLU}{\alpha} (\mathcal{W}_1 \mathcal{Z})\bigr) \Bigr)
\end{equation}

Here, $\mathcal{W}_1 \in \mathbb{R}^{C_f \times \frac{C_f}{\gamma}}$, $\mathcal{W}_2 \in \mathbb{R}^{\frac{C_f}{\gamma} \times C_f}$ denotes the weight matrices of the first and second fully connected layers, and $\gamma$ determines the dimensionality reduction factor controlling the compression ratio. We empirically found that $\gamma=4$ yields optimal performance. Also, $\underset{Sigmoid}{\sigma}$ and $\underset{ReLU}{\alpha}(\cdot)$ represent the sigmoid function and ReLU operator. To retain the non-linear mapping capabilities of the sigmoid function, a dropout layer with a 0.1 probability of dropping units is employed before the sigmoid function. The final step involves rescaling the original feature channels $\mathcal{F}_{\oplus}$ with the earned modulation weights $\mathcal{A}$, as given in~\cref{eq:ccrscale}. This adaptive scaling emphasizes informative channels and suppresses less relevant ones.

\begin{equation}
\label{eq:ccrscale}
f_{CCR}^j=a_k \odot z_k
\end{equation}

The CCR module generates the final output $\mathcal{F}_{CCR}=[f_{CCR}^1,f_{CCR}^2, \ldots , f_{CCR}^{C_f}]$, a tensor containing rescaled feature maps. Each element $f_{CCR}^i$ in $\mathcal{F}_{CCR}$ is obtained by adaptively modulating the corresponding channel $f_{\oplus}^k\in \mathbb{R}^{H \times W \times C_f}$ of the input feature map using the learned activation $a_k \in \mathcal{A}$. This channel-wise modulation, denoted by $\odot$, effectively amplifies informative channels while suppressing less relevant ones. As the network trains, it dynamically tunes these activations, prioritizing features that contribute most to the model's representation power.

\hypertarget{dca}{\paragraph[]{(iii) Dynamic Contextual Attention Module:\eatpunct}}
The Dynamic Contextual Attention (DCA) module employs a novel Transformer-inspired building block, the Multi-Head Dynamic Contextual Attention (MHDCA). MHDCA enhances the representational properties of deep networks by leveraging contextual information among input keys and facilitating self-attention learning. The primary objective of the DCA module is to capture extensive dependencies within the adaptive features ($\mathcal{F}_E^5$) derived from a progressive encoder block while preserving the integrity of spatial information. We utilize a transformer block that includes a Fully Connected (FC) layer, MHDCA, feed-forward layers, and a dropout layer to achieve this. Given the crucial role of the attention mechanism in the DCA module, we provide a detailed introduction to MHDCA before delving into the remaining components.

\noindent\textbf{Multi-Head Dynamic Contextual Attention (MHDCA):} The Multi-Head Dynamic Contextual Attention (MHDCA) module, inspired by the Transformer's success in capturing long-range relationships in sentences, enhances the model's representational power in two key ways. Firstly, MHDCA extends the model's capacity to focus on diverse positions, as the encoding of each head incorporates knowledge about the encodings of the other heads. Secondly, the partitioning of input features broadens the representation subspaces, leading to more indicative attention weights for each partition. Concatenating these enriched representations yields a superior overall representation, ultimately boosting classification accuracy. Within the Dynamic Contextual Attention (DCA) module, MHDCA leverages dynamic convolutions to adapt the parameters of its convolutional kernels dynamically based on specific input samples, further enhancing its representational flexibility.

\begin{figure}[!ht]
    \centering    
    \includegraphics[width=0.5\textwidth]{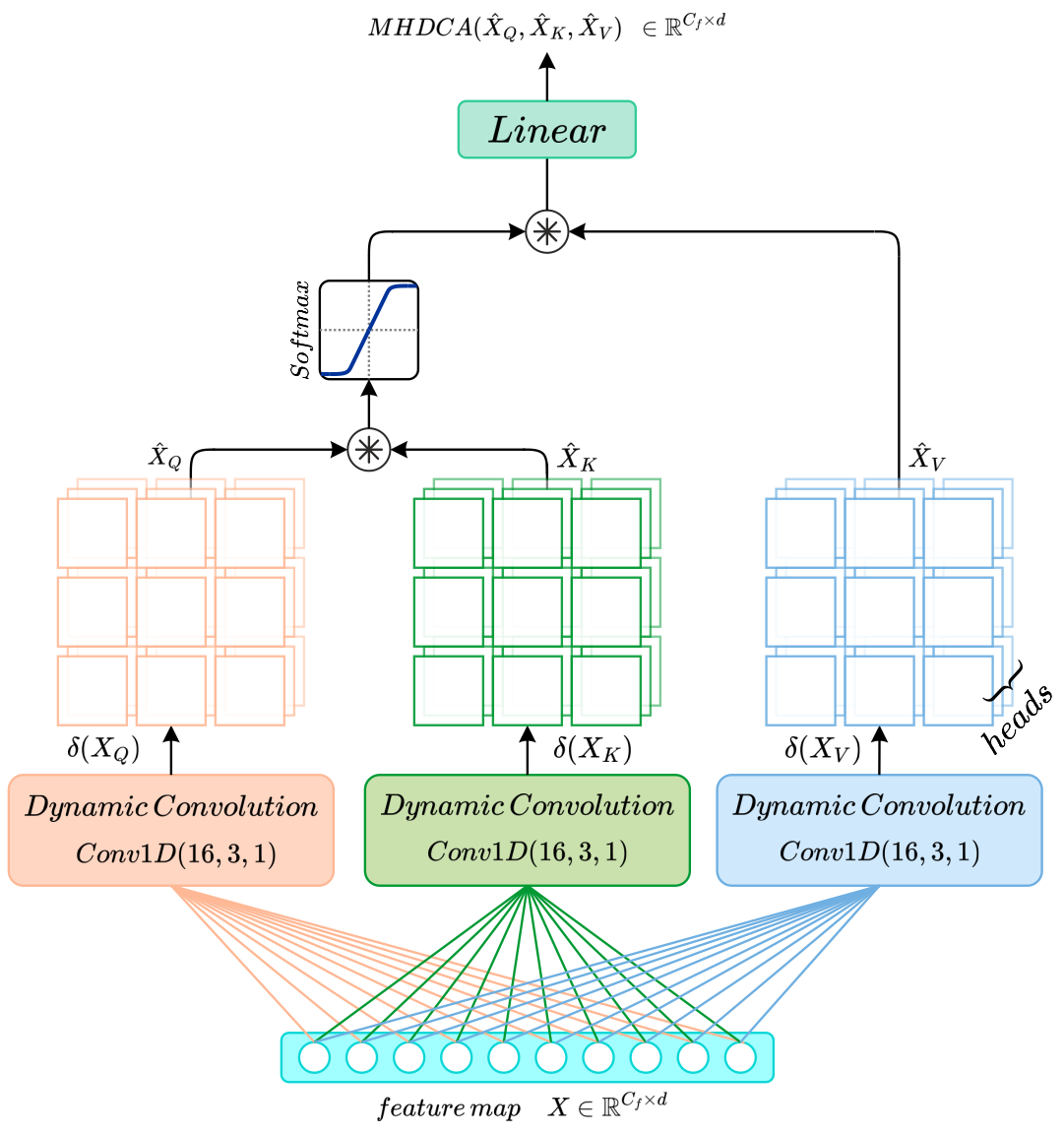}
    \caption {The Architecture of Multi-Head Dynamic Contextual Attention module.}
    \label{fig:tmpednet_mhdca}
\end{figure}

The FC layer of the DCA module outputs $\mathcal{F}^{{FC}_1}$, a tensor of dimension $C_f \times d$, where $C_f$ signifies the number of convolution filters and $d$ represents the length of the tensor. This key input, visualized in~\autoref{fig:tmpednet_mhdca}, feeds into the MHDCA block, extracting the discriminative features. Notably, MHDCA requires three identical copies of $\mathcal{F}^{{FC}_1}$ as inputs. First, it dynamically adjusts the internal representation of $\mathcal{F}^{{FC}_1}$ through a dynamic convolution (represented by $\Omega$), generating $\mathcal{\widehat{F}}^{{FC}_1}=\Omega (\mathcal{F}^{{FC}_1})$. Subsequently, these three transformed matrices, query, key, and value of $\mathcal{\widehat{F}}^{{FC}_1}$ are fed into the attention mechanism described in~\cref{eq:mhdcasoftmax}

\begin{equation}
\label{eq:mhdcasoftmax}
ATT(\mathcal{\widehat{F}}^{{FC}_1}_{\mathcal{Q}},\mathcal{\widehat{F}}^{{FC}_1}_{\mathcal{K}}, \mathcal{\widehat{F}}^{{FC}_1}_{\mathcal{V}})=\underset{Softmax}{\sigma} \left(\frac{(\mathcal{\widehat{F}}^{{FC}_1}_{\mathcal{Q}}) (\mathcal{\widehat{F}}^{{FC}_1}_{\mathcal{K}})^T}{\sqrt{d}} \right) \cdot \mathcal{\widehat{F}}^{{FC}_1}_{\mathcal{V}}
\end{equation}

Here, $(\cdot)$ is the multiplication operation, and $\underset{Softmax}{\sigma}$ represents the softmax function.

We expand the attention mechanism over $\mathcal{H}$ heads for each of the three duplicated matrices derived from the FC layer output to enrich the representation further. Specifically, we partitioned each matrix $\mathcal{\widehat{F}}^{{FC}_1}$ into $\mathcal{H}$ sub-spaces, denoted as $\mathcal{\widehat{F}}^{{FC}_1}_1, \ldots , \mathcal{\widehat{F}}^{{FC}_1}_{\mathcal{H}}$, where $\mathcal{\widehat{F}}^{{FC}_1}_h \in \mathbb{R}^{C_f \times \frac{d}{\mathcal{H}}}$, $1\leq h \leq \mathcal{H}$. Within particular subspaces $h$, an attention similarity score $A^h$ is computed using the equation presented in~\cref{eq:mhdcatt}.

\begin{equation}
\label{eq:mhdcatt}
A^h = ATT(\mathcal{\widehat{F}}^{{FC}_1}_{\mathcal{Q}_h},\mathcal{\widehat{F}}^{{FC}_1}_{\mathcal{K}_h}, \mathcal{\widehat{F}}^{{FC}_1}_{\mathcal{V}_h})\in \mathbb{R}^{C_f \times \frac{d}{\mathcal{H}}}
\end{equation}

Concatenating all $\mathcal{H}$ representations together produces the final result as follows:

\begin{equation}
\label{eq:mhdca}
\mathcal{F}_{MHDCA}=MHDCA(\mathcal{\widehat{F}}^{{FC}_1}_{\mathcal{Q}},\mathcal{\widehat{F}}^{{FC}_1}_{\mathcal{K}}, \mathcal{\widehat{F}}^{{FC}_1}_{\mathcal{V}})=Concat (A^1, \ldots, A^{\mathcal{H}}) 
\end{equation}

Following the MHDCA layer, the DCA module employs a dedicated Feed-Forward (FF) neural network for further feature refinement. This network utilizes a stacked configuration of two FC layers, where each layer leverages the non-linear ReLU activation function to introduce non-linearity, enabling the network to learn complex relationships within the data. Additionally, the FF network incorporates an L1/L2 regularizer to prevent excessive weight values and achieve robust generalization. Formally, this operation can be expressed as $\mathcal{F}_{DCA}= 
\psi_{{FC}_4} \Bigl( r_{{L_1}{L_2}} \bigl(\underset{ReLU}{\sigma} \psi_{{FC}_3} (\mathcal{F}_{MHDCA}) \bigr)\Bigr)$, where $\psi_{FC_{3}}$ and $\psi_{FC_{4}}$ represent the two FC layers, $\underset{ReLU}{\sigma}$ represents the ReLU activation function, and $r_{{L_1}{L_2}}$ represent the L1/L2 regularizer in the DCA module. The resulting $\mathcal{F}_{DCA}$ encapsulates a wider spectrum of contextual relationships, effectively capturing both local and long-range dependencies within the input data by preserving spatial information.

\hypertarget{msas}{\paragraph[]{(iv) Multi-Scale Atrous Spatial Module:\eatpunct}}
In computer vision tasks, extracting detailed features while maintaining global context presents a significant challenge. Balancing feature resolution and receptive field size is crucial, as larger receptive fields capture broader context but sacrifice detail, while smaller ones preserve fine-grained information but lack context. We address this challenge by proposing a novel Multi-Scale Atrous Spatial (MSAS) module that efficiently captures fine and coarse-grained features.

\begin{figure}[!ht]
    \centering    
    \includegraphics[width=0.5\textwidth]{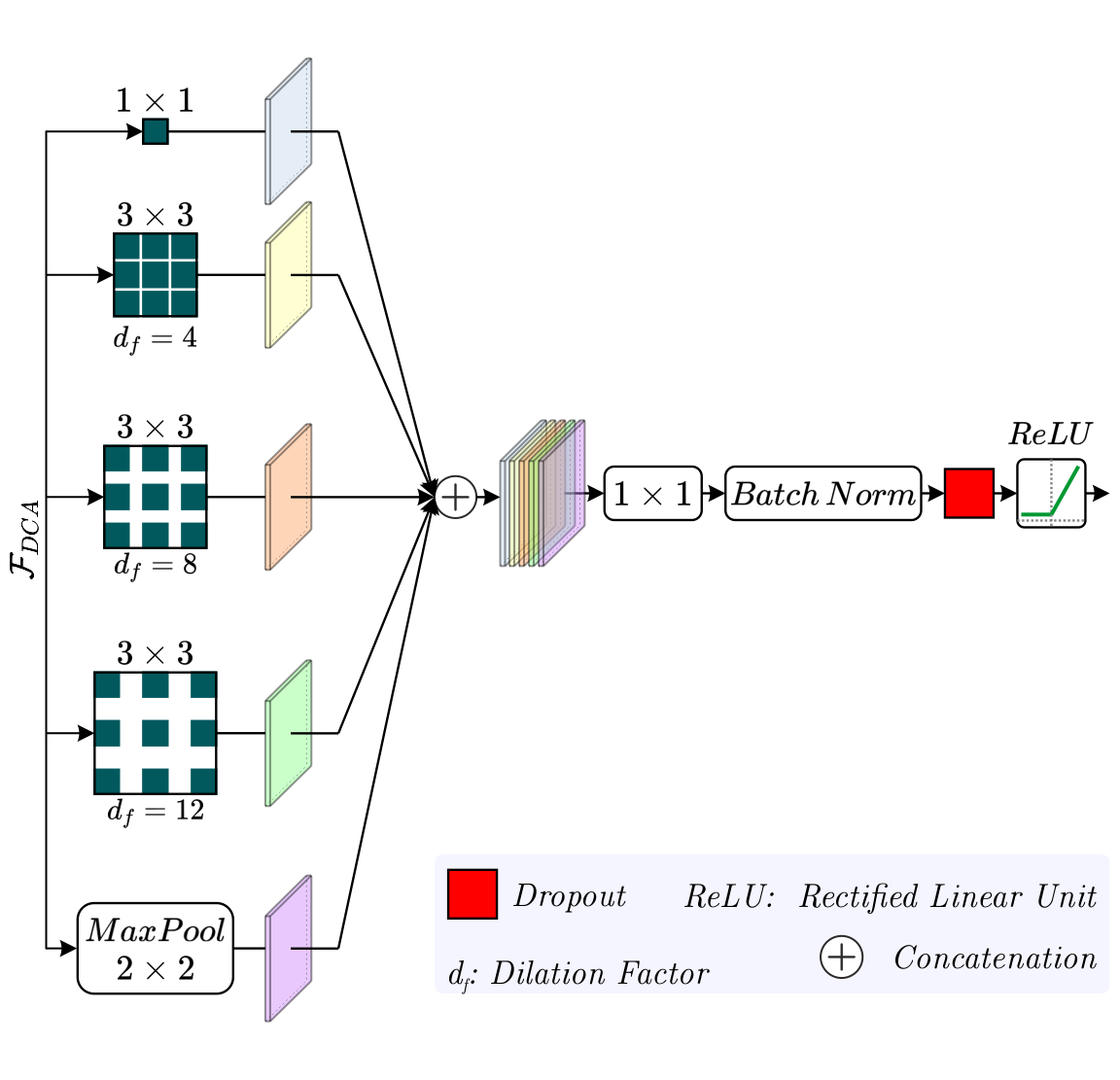}
    \caption {The Architecture of the Multi-Scale Atrous Spatial module.}
    \label{fig:tmpednet_msas}
\end{figure}

MSAS leverages atrous depthwise convolutions followed by pointwise convolutions $(i.e., 1 \times 1$ kernel) to capture multi-scale features from diverse receptive fields with minimal computational overhead. The key lies in utilizing multiple parallel atrous depthwise convolution filters with varying dilation rates. This parallel approach enables MSAS to extract multi-scale features from appropriate receptive fields, capturing details at different scales. These extracted features, representing information from different receptive field sizes, are then concatenated, as shown in~\autoref{fig:tmpednet_msas}. This approach effectively balances the trade-off between feature resolution and receptive field size, allowing MSAS to extract comprehensive contextual information while maintaining computational efficiency.

In~\autoref{fig:tmpednet_msas}, $\mathcal{F}_{DCA}$ denotes the input feature map comprising contextual features from the previous Dynamic Contextual Attention (DCA) module. The output, represented by $\mathcal{F}_{MSAS}$ comprises multi-scale spatial contextual features. The atrous depthwise convolution utilized within the MSAS module is formally defined in~\cref{eq:atconv}:

\begin{equation}
\label{eq:atconv}
\mathcal{F}_{K,d_f}[l]=\sum_{\kappa=1}^{K} \mathcal{F}_{DCA}[l+d_f \ast \kappa] \ast \omega_{\kappa}
\end{equation}

Here, we define $l$ as the location on the feature map, $d_f$ as the dilation factor, $\omega_{\kappa}$ as the $\kappa-$th parameter of the convolution filter, and $K$ as the filter size. As illustrated in~\cref{eq:atconv}, varying the dilation rate $(d_f)$ enables the acquisition of diverse receptive fields, effectively serving as multiple scales for analyzing local spatial components. Specifically, we employ four parallel dilated convolutions with dilation rates of 1, 4, 8, and 12 to extract features at a distinct scale, capturing local details within multiple spatial extents. By concatenating these multi-scale features with distinct receptive fields, MSAS generates high-level contextual features. The mathematical formulation of MSAS is as follows: 

\begin{equation}
\label{eq:msas}
\mathcal{F}_{MSAS}=\mathcal{F}_{3,1} \langle\mathcal{F}_{DCA}\rangle + \mathcal{F}_{3,4} \langle\mathcal{F}_{DCA}\rangle + \mathcal{F}_{3,8} \langle\mathcal{F}_{DCA}\rangle + \mathcal{F}_{3,12} \langle\mathcal{F}_{DCA}\rangle
\end{equation}

Leveraging the multi-scale contextual feature $\mathcal{F}_{MSAS}$, derived from ~\cref{eq:msas}, we progressively upsample and concatenate it with recalibrated features at the corresponding location, feeding this enriched representation into the decoder. Integrating multi-scale information enhances high-level and low-level features with fine-grained semantic information through a simple heuristic approach, improving segmentation performance.

\subsubsection{Morphological Boundary Refinement Module}
\label{sec:boundryrefine}
Segmenting liver and tumor regions presents a challenge due to indistinct boundaries with neighboring organs, resulting in weaker boundary pixels. This limitation is addressed by incorporating edge priors, which provide precise boundary information, significantly improving segmentation accuracy by guiding the model to capture these finer details. The Morphological Boundary Refinement (MBR) module, integrated after the Progressive Encoder-Decoder architecture, enhances the focus on liver and tumor boundaries by actively extracting additional boundary information, leading to more accurate segmentation.

MBR, a post-processing module, leverages 2D morphological erosion to improve the accuracy of liver and tumor segmentation in CT scans. Morphological erosion iteratively shrinks objects in the image iteratively by removing pixels from their edges. Consequently, it selectively eliminates pixels with weak intensities around the boundaries, resulting in a more robust pixel representation with stronger intensities, as mathematically represented in~\cref{eq:mbr}.

\begin{equation}
\label{eq:mbr}
B_{Mask}^{(p,q)}=\begin{cases}
    1, & I_{Mask} \subseteq X_{Mask}^{(p,q)} \\
    0, & \text{\textit{otherwise}}
  \end{cases}
\end{equation}

Here $(p,q)$ locates a specific pixel in $B_{Mask}$. Additionally, $X_{Mask}^{(p,q)}$ 
identifies a structural element from $X_{Mask}$ centered at $(p,q)$. Importantly, the relationship between $I_{Mask}$ and $X_{Mask}$, demonstrating that $I_{Mask}$ falls entirely within the domain of $X_{Mask}$.

\begin{figure*}[!ht]
  \centering
  \includegraphics[width=\textwidth]{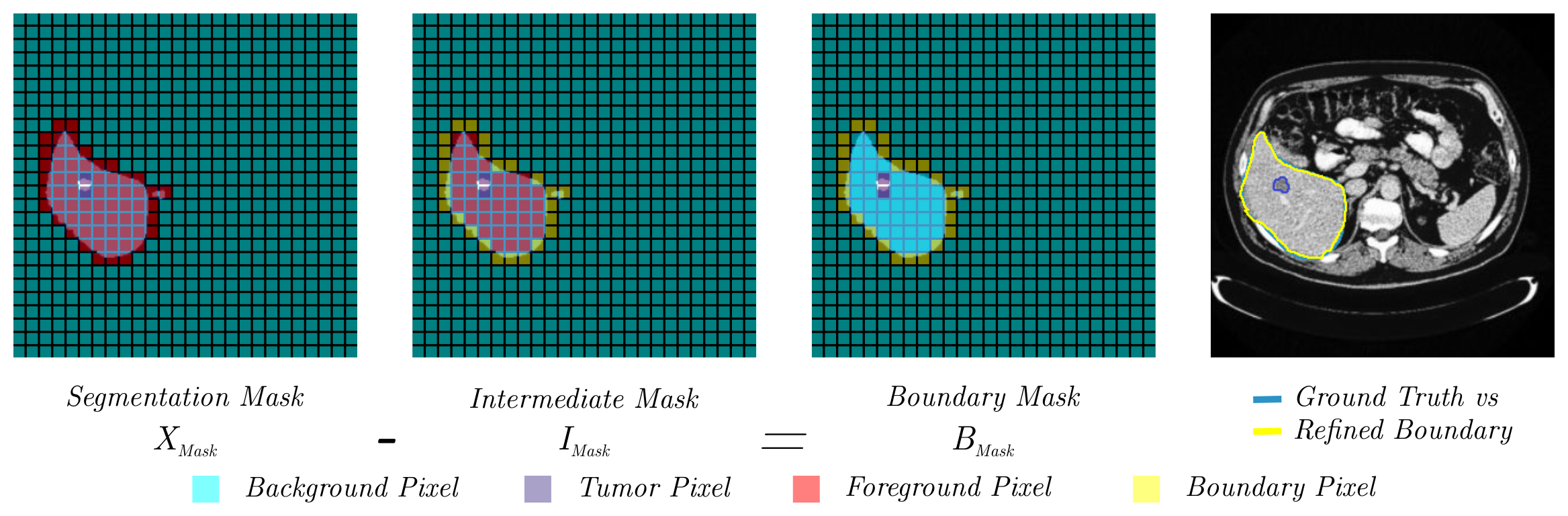}
  \caption{Illustration of Morphological Boundary Refinement (MBR) module for refining the final boundary mask of liver and tumor boundary. The blue color represents the background pixel of the segmentation mask $X_{Mask}$, red represents the foreground pixels, and purple represents the tumor mask. After applying morphological operation, the boundary mask $B_{Mask}$, is marked as light yellow.}
  \label{fig:mbr}
\end{figure*}

As illustrated in~\autoref{fig:mbr}, the MBR module utilizes the input mask $X_{Mask}$ to perform a 2D erosion operation, generating an intermediate mask $I_{Mask}$ that captures the liver and tumor position. Subtracting this $I_{Mask}$ from the $X_{Mask}$ yields the final boundary mask $B_{Mask}$, highlighting the pixels removed during erosion and representing the refined boundary regions. \autoref{fig:mbr} illustrates the process visually, where blue pixels represent the background, red pixels represent the foreground, and light purple pixels denote tumor pixels, primarily located in the refined boundary region. This refined boundary information is then incorporated to improve the overall accuracy of the segmentation process directly.

\autoref{table:param} display the overall configuration of the proposed T-MPEDNet in terms of feature dimensions along with internal layers configuration, respectively. Here, CCR denotes the Compressive Channel Recalibration module.

\begin{table}[!ht]
\caption{Architectural blueprint of the Progressive Encoder-Decoder Network in T-MPEDNet, highlighting feature dimensions and network layer configurations.}
    \centering
    \resizebox{\textwidth}{!}{
    \begin{tabular}{lcccc}
    
        \toprule
        \textbf{{T-MPEDNet Module}}
        & \textbf{Input Dimensions}
        & \textbf{Output Dimensions}
        & \textbf{Network Layer}
        & \textbf{CCR} \\
        
        \midrule
        \multicolumn{5}{l}{\textbf{Progressive Encoder Block}} \\
        
        \midrule
        $AdaFEx_1$ &$256 \times 256 \times 1$&$256 \times 256 \times 16$&Conv2D + Conv2D + Drop + Conv2D& $\times$ \\
        Down-Sampler& $256 \times 256 \times 16$& $128\times 128\times 16$& Max Pooling&\\

        \midrule
        $AdaFEx_2$ &$128 \times 128 \times 16$&$128 \times 128 \times 32$&Conv2D + Conv2D + Drop + Conv2D& $\times$ \\
        Down-Sampler& $128 \times 128 \times 32$& $64\times 64\times 32$& Max Pooling&\\

        \midrule
        $AdaFEx_3$ &$64\times 64\times 32$&$64\times 64\times 64$&Conv2D + Conv2D + Drop + Conv2D& $\times$ \\
        Down-Sampler&$64\times 64\times 64$&$32\times 32\times 64$& Max Pooling&\\

        \midrule
        $AdaFEx_4$ &$32\times 32\times 64$&$32\times 32\times 128$&Conv2D + Conv2D + Drop + Conv2D& $\times$ \\
        Down-Sampler&$32\times 32\times 128$&$16\times 16\times 128$& Max Pooling&\\

        \midrule
        $AdaFEx_5$ &$16\times 16\times 128$&$16\times 16\times 256$&Conv2D + Conv2D + Drop + Conv2D& $\times$ \\
        Down-Sampler&$16\times 16\times 256$&$8\times 8\times 256$& Max Pooling&\\ 

        \midrule
        \multicolumn{5}{l}{\textbf{Dynamic Contextual Attention Module}} \\
        \midrule
        DCA &$8\times 8\times 256$&$8\times 8\times 256$&FC+MHDCA+FC+Drop& $\times$ \\

        \midrule
        \multicolumn{5}{l}{\textbf{Multi-Scale Atrous Spatial Module}} \\
        \midrule
        MSAS &$8\times 8\times 256$&$8\times 8\times 512$&$d_f(4\oplus8\oplus12)$ + BN + Drop & $\times$ \\

        \midrule
        \multicolumn{5}{l}{\textbf{Progressive Decoder Block}} \\
        
        \midrule
        Up-Sampler&$8\times 8\times 512$&$16\times 16\times 256$& Conv2DTranspose& \checkmark \\
        $AdaFEx_5$&$16\times 16\times 256$&$16\times 16\times 256$&Conv2D + Conv2D + Drop + Conv2D&\\

        \midrule
        Up-Sampler &$16\times 16\times 256$&$32\times 32\times 128$& Conv2DTranspose& \checkmark \\
        $AdaFEx_4$&$32\times 32\times 128$&$32\times 32\times 128$&Conv2D + Conv2D + Drop + Conv2D&\\

        \midrule
        Up-Sampler &$32\times 32\times 128$&$64\times 64\times 64$&Conv2DTranspose& \checkmark \\
        $AdaFEx_3$&$64\times 64\times 64$&$64\times 64\times 64$&Conv2D + Conv2D + Drop + Conv2D&\\

        \midrule
        Up-Sampler &$64\times 64\times 64$&$128\times 128\times 32$&Conv2DTranspose& \checkmark \\
        $AdaFEx_2$&$128\times 128\times 32$&$128\times 128\times 32$&Conv2D + Conv2D + Drop + Conv2D&\\

        \midrule
        Up-Sampler &$128\times 128\times 32$&$256 \times 256 \times 16$&Conv2DTranspose& \checkmark \\
        $AdaFEx_1$& $256 \times 256 \times 16$& $256\times 256\times 1$&Conv2D + Conv2D + Drop + Conv2D&\\

        \bottomrule
        \multicolumn{5}{l}{\scriptsize CCR included: \checkmark;  CCR not-included: $\times$.}
    \end{tabular}
    } 
\label{table:param}
\end{table}

\section{Experiments and Results}
\label{sec:experiment}
In this section, we present the particulars of the diverse datasets used in this research work for experimental evaluation, which is further compared with various baseline methods. We subsequently benchmark our proposed model against established baselines, utilizing rigorous evaluation metrics to highlight its superior performance and overall effectiveness. 

\subsection{Experimental Setup}
\subsubsection{Dataset Summarization}
The proposed T-MPEDNet framework was tested using two publicly available datasets, namely, Liver Tumor Segmentation Benchmark(LiTS)~\citep{Bilic2023lits} and the 3D Image Reconstruction for Comparison of Algorithm Database (3DIRCADb)~\citep{soler20103d}. \autoref{fig:dataset} depicts a selection of CT scans utilized for the quantitative and qualitative analysis of the proposed T-MPEDNet framework. To achieve a comprehensive assessment of T-MPEDNet's generalizability, we randomly split CT images into 80\% for training, 10\% for validation, and 10\% for independent testing, as presented in ~\autoref{table:dataset}. We briefly summarize the dataset as follows:

\begin{figure}[!ht]
  \centering
  \includegraphics[width=0.65\textwidth]{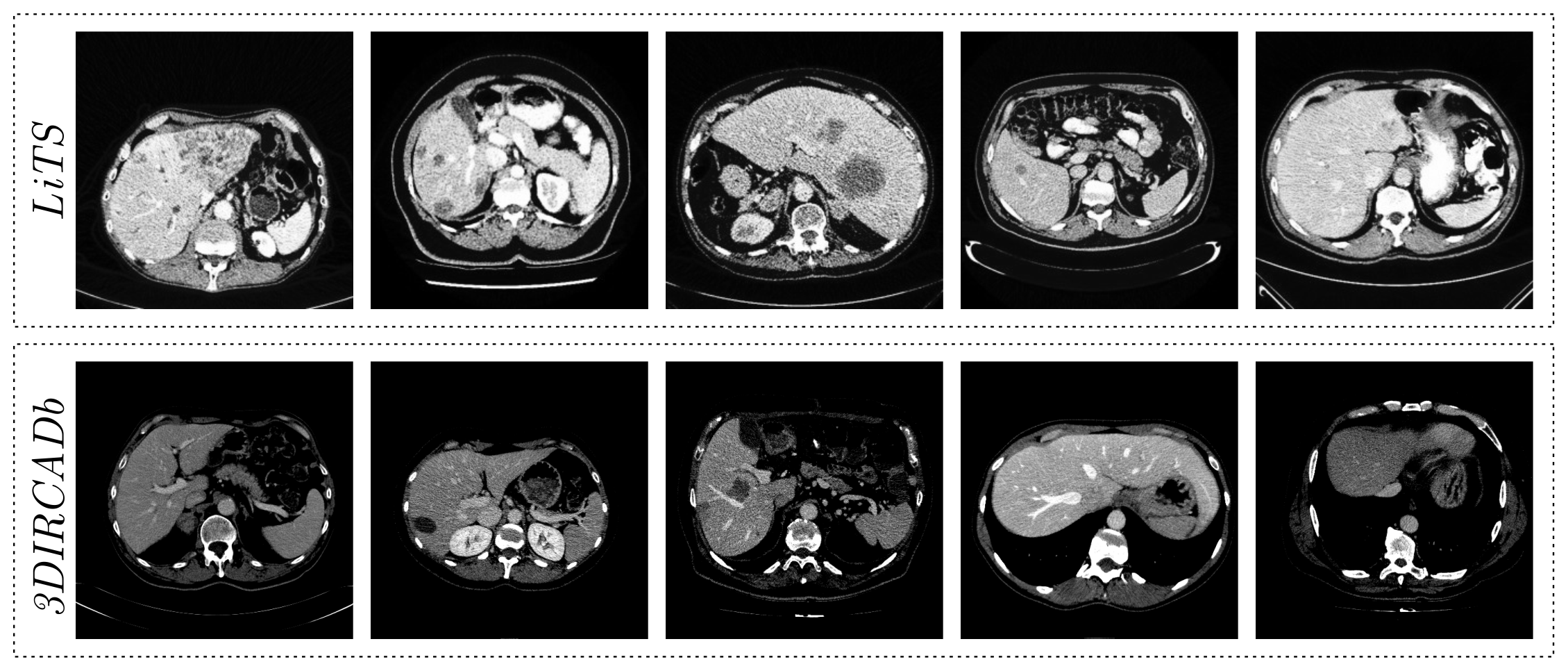}
  \caption{Axial CT slices from the LiTS and 3DIRCADb datasets, highlighting the variety of liver and tumor anatomies present in the study.}
  \label{fig:dataset}
\end{figure}

\paragraph[]{LiTS: \eatpunct}
The Liver Tumor Segmentation Challenge (LiTS) dataset~\citep{Bilic2023lits} is a popular benchmark for evaluating automated liver and tumor segmentation techniques in contrast-enhanced abdominal CT scans. Launched in conjunction with ISBI 2017 and MICCAI 2017, LiTS comprises 131 anonymized labeled CT scans from various clinical sites worldwide. In a single volume, the slice count exhibits significant variability, ranging from 75 to 841. The voxel spacing within the scans varies along the X, Y, and Z axes, with values spanning from 0.6 to 0.98 mm, 0.6 to 0.96 mm, and 0.4 to 0.45 mm, respectively. \autoref{table:dataset} presents a detailed overview of the dataset characteristics.

\paragraph[]{3DIRCADb: \eatpunct}
The 3DIRCADb dataset~\citep{soler20103d} features 20 anonymous CT volumes (10 male, 10 female) acquired across European hospitals from different patients. These volumes were manually annotated by expert radiologists to highlight various structures of interest. 75 percent of patients possess tumors with varying sizes, shapes, and locations. The remaining 25 percent comprises non-tumor reference cases, essential for assessing the segmentation's generalizability across diverse clinical techniques. Encompassing a wide range of tumor characteristics within the dataset allows for a comprehensive assessment of the segmentation technique's robustness and generalizability. The single volume of CT scan Scans exhibits non-uniform voxel spacing, with values ranging from 0.56-0.87 mm in the X axis, 0.56 - 0.87 mm in the Y axis, and 01.6 - 4.05 mm in the Z axis. \autoref{table:dataset} comprehensively summarizes the key characteristics of the dataset.

\begin{table*}[!ht] 
\centering
\caption{Description of the LiTS and 3DIRCADb public datasets utilized for liver and tumor segmentation evaluation.}
    \resizebox{\textwidth}{!}{
    \begin{tabular}{llcccccc}
    \toprule
        & \textbf{Dataset} & \textbf{Number} & \textbf{Spatial} & \textbf{Slices in} & \textbf{X-axis Voxel} & \textbf{Y-axis Voxel} & \textbf{Z-axis Voxel} \\

        & & \textbf{of slices in} & \textbf{Resolution} & \textbf{the volumes } & \textbf{spacing} & \textbf{spacing} & \textbf{spacing} \\

        & & \textbf{CT Samples} & \textbf{of each slice} & \textbf{(in mm)} & \textbf{(in mm)} & \textbf{(in mm)} & \textbf{(in mm)} \\
    \midrule

    & & & & & & &  \\  [-5pt]
    \multirow{2}{*}{\rotatebox[origin=c]{90}{LiTS2017}}
    & Training & 54065 
    & \multirow{4}{*}{$512 \times 512$} & \multirow{4}{*}{75-841} 
    & \multirow{4}{*}{0.60 - 0.98} 
    & \multirow{4}{*}{0.60 - 0.98} 
    & \multirow{4}{*}{0.45 - 0.50} \\
    & & & & & & & \\  [-5pt]
    & Validation & 5001 & & & & &  \\ 
    & & & & & & &  \\  [-5pt]    
    & Testing & 4573 & & & & & \\
    & & & & & & & \\  [-5pt]
    \midrule

    & & & & & & &  \\  [-5pt]
    \multirow{2}{*}{\rotatebox[origin=c]{90}{3DIRCADb}}
    & Training & 2258 
    & \multirow{4}{*}{$512 \times 512$} & \multirow{4}{*}{74-260} 
    & \multirow{4}{*}{0.56-0.87} 
    & \multirow{4}{*}{0.56-0.87} 
    & \multirow{4}{*}{1.60-4.00}  \\
    & & & & & & & \\  [-5pt]
    & Validation & 282 & & & & &  \\ 
    & & & & & & &  \\  [-5pt]
    & Testing & 282 & & & & & \\
    & & & & & & & \\  [-5pt]
    \bottomrule
    \end{tabular}
    } 
\label{table:dataset}
\end{table*}

\subsubsection{Comparison Methods}
This section provides a concise overview of the existing methods, highlighting their key techniques and functionalities as a basis for comparison with the proposed method.

\textbf{Encoder Decoder-based Segmentation:} The existing baseline models for automated liver and tumor segmentation utilize encoder-decoder architectures are explored. Liao et al.~\cite{Liao2024mscfunet} employ a multi-scale context fusion strategy to extract both multi-scale local features and long strip features. The integrated dual self-attention mechanism utilizes one-hot operations within shift convolution, aiming to enhance global information capture in the spatial domain. Slice-Fusion~\citep{Tu2023slice} proposed a novel framework to utilize the global structural relationships between tissues throughout neighboring slices and fuse their features according to tissue importance. DRAUNet~\citep{Chen2023draunet} presented an approach utilizing a Deep Residual (DR) block and a Deep Attention Module (DAM). The method integrated coronal slices with transverse slices to improve the segmentation process. AutoLT~\citep{Di2022autolt} introduced a novel approach for automated liver tumor segmentation that employs hierarchical iterative superpixels and local statistical features.

\textbf{Attention-based Segmentation:} This section explores existing baseline methods that incorporate the attention mechanism into their segmentation strategies. AC-E Net~\citep{Li2023ace} devised an Attentive Context Encoding Module (ACEM) for better representation of 3D features, along with an additional complemental loss to focus on the region and the boundary. HRFU-Net~\citep{Kushnure2022hfru} modified the original UNet~\citep{Ronneberger2015unet} to include a feature fusion mechanism and local feature reconstruction. In addition, the Atrous Spatial Pyramid Pooling (ASPP) module is also placed in the bottleneck of the architecture. LITS-Net~\citep{Li2022litsnet} outlined the Shift-Channel Attention Module (S-CAM) to model feature interdependencies. TD-Net~\citep{Di2023tdnet} proposed a model where direction information and Transformer mechanism were incorporated in the Convolution Mechanism.

\textbf{Multi-Scale Feature Extractor-based Segmentation:} Focusing on automated liver and tumor segmentation, we examine existing baseline approaches that leverage multi-scale feature extraction. MAUNet~\citep{Jiang2023rmau} suggested novel modules that utilized residual connections and inter-spatial information. This was achieved by exploiting rich multi-scale feature information and capturing interchannel relationships to improve accuracy. FRAUNet~\citep{Chen2022fraunet} demonstrated an automated dual-step segmentation network utilizing a cascading framework for initial segmentation and further a fully connected Conditional Random Field (CRF) to refine the results. KiU-Net~\citep{Valanarasu2022kiunet} constructed a dual branch model that focused on high-level features through a traditional UNet and low-level features through a convolutional architecture which is overcomplete. DefED-Net~\citep{Lei2022defednet} leverages deformable convolution to improve feature representation capabilities and introduces a novel Ladder-Atrous-Spatial-Pyramid-Pooling (Ladder-ASPP) module.

\subsubsection{Evaluation Metrics}
We use established metrics to quantitatively evaluate the model's performance, with the Dice Similarity Coefficient (DSC) precisely measuring the overlap between predicted segmentation and the corresponding ground truth masks. The averaged Dice score for each category is reported for transparency and fairness in comparison. The Dice similarity coefficient is mathematically defined as:

\begin{equation}
\label{eq:emdice}
DSC = \frac{2 \times |CT_{P} \cap CT_G|}{|CT_{P}| + |CT_G|}
\end{equation}

Here, $CT_{P}$ is the set of elements in the predicted segmentation mask, and $CT_{G}$ is the ground truth segmentation mask. $\cap$ denotes the common pixels between the $CT_{P}$ and $CT_{G}$.

\subsubsection{Implementation Details}
We conducted our experiments using Python and the TensorFlow library on an NVIDIA P100 GPU with 16GB of GPU RAM and 512 GB of system RAM. To optimize the model's trainable parameters, we employed the Adam optimizer with an initial learning rate of 1e-5 and implemented an adaptive learning rate strategy. This dynamic learning rate strategy utilizes a callback function to monitor validation accuracy. If validation accuracy remains unchanged for a specified number of training steps, the callback function automatically reduces the learning rate by 0.65. To counter overfitting, we incorporated Keras image augmentation techniques. These techniques randomly rotated, flipped, and zoomed the input data, creating diverse training samples that enhanced the model's generalizability to unseen examples. This augmented data improved the model's robustness and minimized the risk of memorizing specific training patterns. We utilized both L1 and L2 regularization during training to encourage sparsity and promote network stability. L1 regularization enforced a Laplace before the weight parameters, promoting sparsity and encouraging the model to rely on a select subset of the most influential features. L2 regularization, on the other hand, served as a constraint on the weight magnitudes, preventing overfitting and improving model stability.

\subsection{Result and Analysis}
This section undertakes a comprehensive evaluation of the proposed T-MPEDNet against state-of-the-art methods. We further conduct targeted ablation studies, systematically isolating and evaluating the contribution of each module within the framework, thereby pinpointing their impact and efficacy.

\subsubsection{Comparison with State-of-the-Art Methods}
In this section, we compare the performance of our T-MPEDNet with twelve state-of-the-art (SOTA) liver and tumor segmentation methods, including MSCF-Net~\citep{Liao2024mscfunet}, Slice-Fusion~\citep{Tu2023slice}, DRAUNet~\citep{Chen2023draunet}, AutoLT~\citep{Di2022autolt}, AC-E Net~\citep{Li2023ace}, HFRU-Net~\citep{Kushnure2022hfru}, LiTS-Net~\citep{Li2022litsnet}, TD-Net~\cite{Di2023tdnet}, RMAU-Net~\citep{Jiang2023rmau}, FRA-UNet~\citep{Chen2022fraunet}, KiU-Net~\citep{Valanarasu2022kiunet}, and DefED-Net~\citep{Lei2022defednet}. We test learning ability across two diverse datasets, presenting both quantitative and qualitative results to assess its capabilities comprehensively.

\begin{table*}[!ht]
\caption{Quantitative comparison of the proposed T-MPEDNet with state-of-the-art methods for liver and tumor segmentation on LiTS and 3DIRCADb. The Dice Score Coefficient (higher implies better) is reported for each dataset. The best and second-best results are highlighted in \colorbox{inchworm}{green} and \colorbox{yellow}{yellow}, respectively. $|\nabla|$ denotes the absolute performance drop relative to T-MPEDNet.}
    \centering
    \resizebox{0.75\textwidth}{!}{
    \begin{tabular}{lcccccccc}
        \toprule
        \textbf{{Methods}}
        &\multicolumn{4}{c}{\textbf{\thead{LiTS}}} 
        &\multicolumn{4}{c}{\textbf{\thead{3DIRCADb}}} \\
        \cmidrule(l){2-5} \cmidrule(l){6-9} 
        
        &\multicolumn{2}{c}{\textbf{\thead{Liver}}}
        &\multicolumn{2}{c}{\textbf{\thead{Tumor}}}
        &\multicolumn{2}{c}{\textbf{\thead{Liver}}}
        &\multicolumn{2}{c}{\textbf{\thead{Tumor}}}\\ [-3pt] 
        
        \cmidrule(l){2-3} \cmidrule(l){4-5} \cmidrule(l){6-7} \cmidrule(l){8-9}
        & \textbf{DSC$^a$}  & $|\nabla|$ & \textbf{DSC$^a$}  & $|\nabla|$& \textbf{DSC$^a$}  & $|\nabla|$& \textbf{DSC$^a$}  & $|\nabla$\\

        \midrule    
        \multicolumn{9}{l}{\textbf{Encoder Decoder-based Segmentation}} \\
        \midrule
        MSCF-Net~\citep{Liao2024mscfunet} & 96.4 & 1.2 & 87.3 & 1.8 & 96.5 & 1.8 & 78.2 & 5.1 \\
        Slice-Fusion~\citep{Tu2023slice} & 95.6 & 2.0 & 86.4 & 2.7 & 95.4 & 2.9 & 77.1 & 6.2 \\
        DRAUNet~\citep{Chen2023draunet} & 96.9 & 0.7 & 86.6 & 2.5 & 97.2 & 1.1 & 77.4 & 5.9 \\
        AutoLT~\citep{Di2022autolt} & 93.8 & 3.8 & 84.9 & 4.2 & 96.3 & 2.0 & 75.7 & 7.6 \\
        
        \midrule
        \multicolumn{9}{l}{\textbf{Attention-based Segmentation}}\\
        \midrule
        AC-E Net~\citep{Li2023ace} & 94.6 & 3.0 & 84.5 & 4.6 & 97.4 & 0.9 & 76.5 & 6.8 \\
        HFRU-Net~\citep{Kushnure2022hfru} & 95.5 & 2.1 & 83.4 & 5.7 & \colorbox{yellow}{97.9} & 0.4 & 79.2 & 4.1 \\
        LiTS-Net~\citep{Li2022litsnet} & 96.0 & 1.6 & 82.1 & 7.0 & 96.7 & 1.6 & 80.1 & 3.2 \\
        TD-Net~\cite{Di2023tdnet} & \colorbox{yellow}{97.1} & 0.5 & 85.2 & 3.9 & 97.8 & 0.5 & 79.4 & 3.9 \\

        \midrule
        \multicolumn{9}{l}{\textbf{Multi-Scale Feature Extractor-based Segmentation}}\\
        \midrule
        RMAU-Net~\citep{Jiang2023rmau} & 95.7 & 1.9 & 83.1 & 6.0 & 97.5 & 0.8 & \colorbox{yellow}{82.1} & 1.2 \\
        FRA-UNet~\citep{Chen2022fraunet} & 96.8 & 0.8 & 85.5 & 3.6 & 97.7 & 0.6 & 75.1 & 8.2 \\
        KiU-Net~\citep{Valanarasu2022kiunet} & 93.5 & 4.1 & 82.6 & 6.5 & 97.1 & 1.2 & 76.3 & 7.0 \\
        DefED-Net~\citep{Lei2022defednet} & 97.0 & 0.6 & \colorbox{yellow}{88.1} & 1.0 & 96.8 & 1.5 & 74.2 & 9.1 \\

        \midrule
        \multicolumn{9}{l}{\textbf{Transformer-aware Multiscale Progressive Encoder-Decoder Network}}\\
        \midrule        
        T-MPEDNet (ours) & \colorbox{inchworm}{97.6} & 0.0 & \colorbox{inchworm}{89.1} & 0.0 & \colorbox{inchworm}{98.3} & 0.0 & \colorbox{inchworm}{83.3} & 0 \\
        \bottomrule
    \end{tabular}
    } 
\label{table:sotalits3d}
\end{table*}

\paragraph[]{A. Quantitative Comparison:\eatpunct}
This study conducts a performance comparison of various liver and tumor segmentation methods on the widely employed LiTS~\citep{Bilic2023lits} and 3DIRCADb~\citep{soler20103d} datasets. We employ the Dice Similarity Coefficient (DSC) to evaluate model performance, quantifying the overlap between predicted and ground truth regions. Higher DSC values correspond to better segmentation results. To facilitate a comprehensive comparison, we categorize existing state-of-the-art (SOTA) methods into three categories: Encoder Decoder-based Segmentation, Attention-based Segmentation, and Multi-Scale Feature Extractor-based Segmentation. The Transformer-aware Multiscale Progressive Encoder-Decoder Network (T-MPEDNet), the proposed framework in the paper, achieved the highest DSC values for both liver and tumor segmentation on both datasets.

T-MPEDNet adopts a transformer-based encoder-decoder architecture equipped with multi-scale feature extraction and fusion modules. This design empowers the model to simultaneously capture global and local context information, leading to accurate and consistent segmentation outputs. From~\autoref{table:sotalits3d}, it is observed that T-MPEDNet achieved a DSC of 97.6 for liver segmentation and 89.1 for tumor segmentation on the LiTS~\citep{Bilic2023lits} dataset, surpassing all other methods by a clear margin. Notably, T-MPEDNet outperforms the second-best method, TD-Net~\citep{Di2023tdnet} by 0.5\% for liver segmentation and DefED-Net~\citep{Lei2022defednet} by 1.0\% for tumor segmentation. Similarly, on the 3DIRCADb~\citep{soler20103d} dataset, T-MPEDNet achieved a DSC of 98.3 for liver segmentation and 83.3 for tumor segmentation, demonstrating its generalizability across diverse datasets. Again, T-MPEDNet outperforms the second-best method, HFRU-Net~\citep{Kushnure2022hfru} by 0.4\% for liver segmentation and RMAU-Net~\citep{Jiang2023rmau} by 1.2\% DSC points for tumor segmentation. These findings are further emphasized by the visual representation of the absolute performance drop ($|\nabla|$) relative to T-MPEDNet. As shown in~\autoref{table:sotalits3d}, all other methods exhibit a clear performance drop compared to T-MPEDNet for both liver and tumor segmentation on both datasets. This visual representation reinforces the quantitative results, demonstrating T-MPEDNet's superiority in liver and tumor segmentation tasks. These results suggest that the transformer-aware multiscale progressive encoder-decoder architecture is a promising approach for liver tumor segmentation.

\begin{figure*}[!ht]
  \centering
  \includegraphics[width=\textwidth]{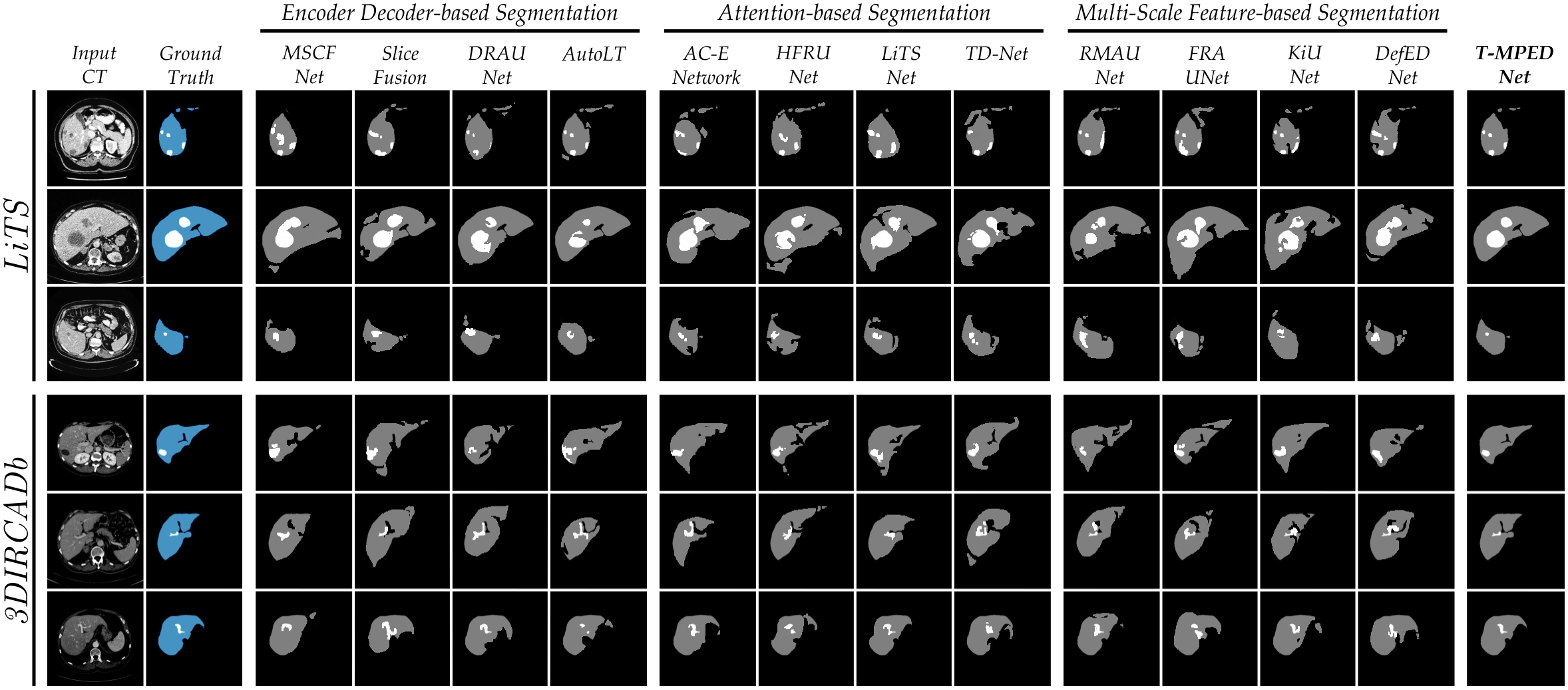}
  \caption{Qualitative assessment of the visual quality of T-MPEDNet segmentation outputs on the LiTS dataset, compared to existing baseline methods.}
  \label{fig:qualsota}
\end{figure*}

\paragraph[]{B. Qualitative Comparison:\eatpunct} \autoref{fig:qualsota} show the qualitative segmentation results of the proposed T-MPEDNet and other competing models on the LiTS~\citep{Bilic2023lits} and 3DIRCADb~\citep{soler20103d} dataset. Compared with other methods, our proposed framework can precisely segment and locate complex liver and small tumors and suppress the non-region of interest (NROI), which includes multiple organs surrounding the liver anatomy. Even for multiple tumors, T-MPEDNet can capture more tiny and complex tumors and is closer to the ground truth.

\autoref{fig:qualsota} shows the visual comparison of the three CT slices on LiTS~\citep{Bilic2023lits} dataset for liver and tumor segmentation. First, from the segmentation results obtained by Encoder-Decoder-based segmentation, DRAUNet~\citep{Chen2023draunet} and AutoLT~\citep{Di2022autolt} show comparable results for liver and tumor segmentation. The results from T-MPEDNet are superior because of the leverage progressive encoder-decoder, which aims to extract deep fine-grained features from CT scans. Secondly, in Attention-based segmentation, T-MPEDNet's segmentation results are almost similar to the ground truth images. Only LiTS-Net~\citep{Li2022litsnet} shows similar performance for liver segmentation but not for tumor segmentation. T-MPEDNet incorporates the transformer-inspired Dynamic Contextual Attention (DCA) module to capture deep global features efficiently and learn contextual relationships among spatial information. DCA module integrates dynamic convolutions before multiheaded attention to map the global and spatial patterns. Lastly, in Multi-Scale Feature Extractor-based segmentation, DefED-Net~\citep{Lei2022defednet} efficiently segments the tumor from complex liver anatomy and leads to the second-best tumor segmentation results. However, T-MPEDNet utilizes multiple levels of atrous convolution to learn local detail features effectively and ensures accurate and reliable predictions.

\autoref{fig:qualsota} shows the visual comparison of the three CT slices on 3DIRCADb~\citep{soler20103d} dataset for liver and tumor segmentation. DRAUNet~\citep{Chen2023draunet} meticulously segments the liver whereas AutoLT~\citep{Di2022autolt} segments tumor inefficiently and unable to identify tumor boundaries under the Encoder-Decoder-based segmentation technique. T-MPEDNet efficiently utilizes the adaptive features from the AdaFEx module and progressively extracts deep context from low-level details to higher-level semantic representations within the Encoder-Decoder block. HFRU-Net~\citep{Kushnure2022hfru} generates better results than all other Attention-based segmentation methods. T-MPEDNet leverages the power of dynamic convolutions, which were added before multiheaded attention. RMAU-Net~\citep{Jiang2023rmau} proficiently segments the tumor and liver with a marginal DSC because they incorporate multi-scale receptor modules in their architecture. T-MPEDNet integrates a Multi-Scale Atrous Spatial module with larger receptive fields to capture broader context while preserving fine-grained information.

The qualitative results presented in \autoref{fig:qualsota} further demonstrate T-MPEDNet's superiority in segmenting both livers and tumors compared to state-of-the-art (SOTA) techniques. Notably, in the fourth column of both figures, T-MPEDNet exhibits focused attention on the liver region, effectively suppressing the influence of surrounding organs while precisely segmenting even tiny tumors, resulting in smoother boundaries compared to competing approaches. These results suggest that T-MPEDNet's ability to learn adaptive features seamlessly integrated with multi-scale contextual feature extraction contributes to its superior segmentation accuracy. 

\subsubsection{Ablation Study}
To assess the contributions of the key modules within the T-MPEDNet framework, we systematically assess the effectiveness of all proposed components and strategies to enhance the overall performance of our framework. The results are shown in~\autoref{table:ablation}. Specifically, we have explored the effects of the Progressive Encoder-Decoder (PED) block, Compressive Channel Recalibration (CCR) module,  Dynamic Contextual Attention (DCA) module, Multi-Scale Atrous Spatial (MSAS) module, and Morphological Boundary Refinement (MBR) module. The key takeaway is that combining all modules (PED, CCR, DCA, MSAS, MBR) achieves the best Dice Similarity Coefficient (DSC) for both liver and tumor of 97.6 and 89.1.

\begin{table*}[!b]
\caption{Ablation Study of proposed T-MPEDNet on the LiTS Dataset.}
\centering
    \resizebox{0.85\textwidth}{!}{
    \begin{tabular}{ccccccccc}
        \toprule
        \textbf{{PED}}
        &\textbf{{CCR}}
        &\textbf{{DCA}}
        &\textbf{{MSAS}}
        &\textbf{{MBR}}
        &\multicolumn{2}{c}{\textbf{\thead{Liver Segmentation}}} 
        &\multicolumn{2}{c}{\textbf{\thead{Tumor Segmentation}}} \\ 
        
        \cmidrule(l){6-7} \cmidrule(l){8-9}            
        & & & & & \textbf{DSC$^a$} & $\nabla$ & \textbf{ DSC$^a$} & $\nabla$ \\

        \midrule
        \checkmark & \checkmark & \checkmark & \checkmark & \checkmark
        & 97.6 & 0.0& 89.1 & 0.0\\
        
        \checkmark & \checkmark & \checkmark & \checkmark & $\times$
        & 97.4 & 0.2& 88.6 & 0.5\\

        \checkmark & \checkmark & \checkmark & $\times$ & $\times$
        & 96.7 & 0.7& 86.2 & 2.9\\

        \checkmark & \checkmark & $\times$ & $\times$ & $\times$
        & 95.4 & 2.2& 83.9 & 5.2\\

        \checkmark & $\times$ & $\times$ & $\times$ & $\times$
        & 93.5 & 4.1 &81.4 & 7.7\\
        \bottomrule

        \multicolumn{9}{l}{\scriptsize PED: Progressive Encoder-Decoder Block, CCR: Compressive Channel Recalibration Module, DCA: Dynamic Contextual }\\
        \multicolumn{9}{l}{\scriptsize Attention Module, MSAS: Multi-Scale Atrous Spatial Module and MBR: Morphological Boundary Refinement Module,}\\
        \multicolumn{9}{l}{\scriptsize DSC$^a$: Dice Similarity Coefficient, $|\nabla|$: Absolute performance drop relative to T-MPEDNet}        
    \end{tabular}
    } 
\label{table:ablation}
\end{table*}

\begin{figure}[!ht]
    \centering    
    \begin{minipage}{0.49\textwidth}
    \centering
    \includegraphics[width=\textwidth]{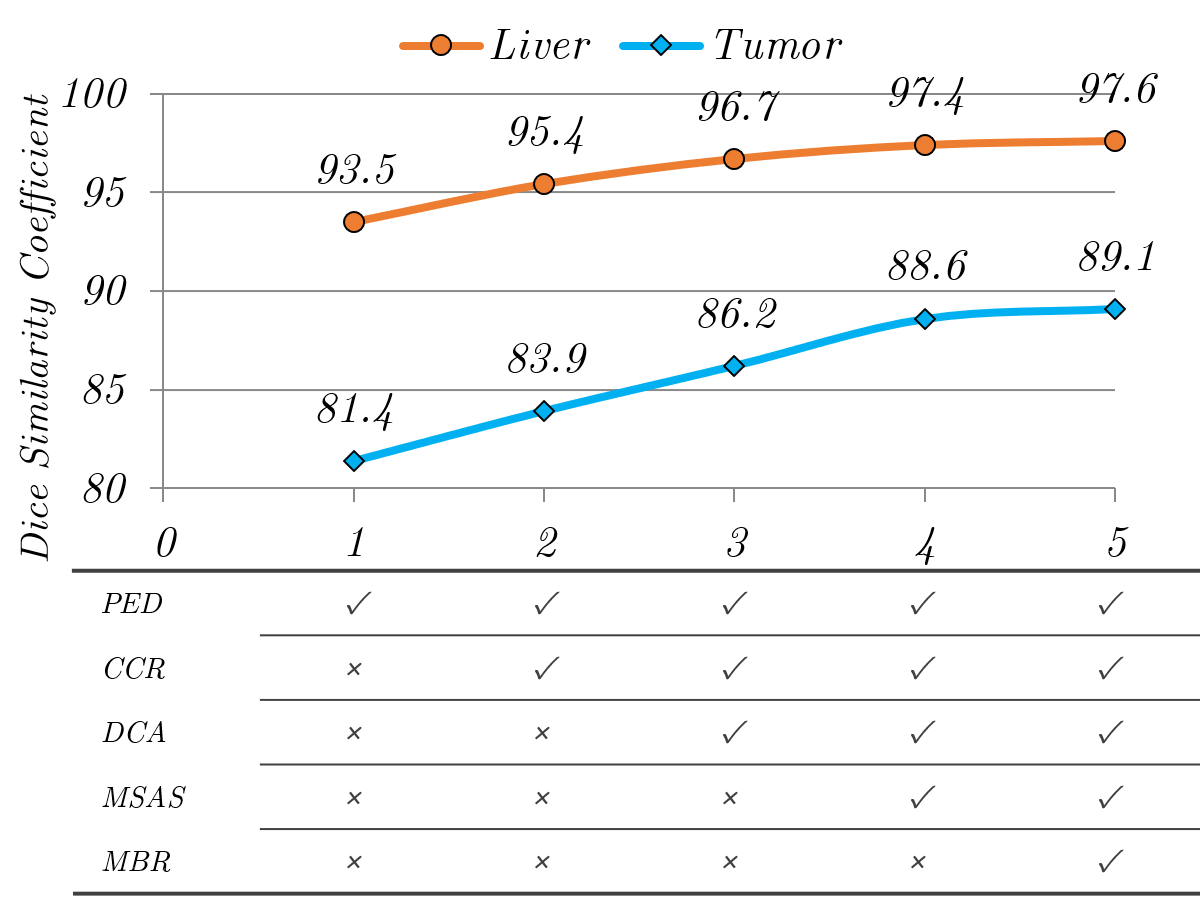}
    \caption {Impact analysis curves for T-MPEDNet components on liver and tumor segmentation performance using Dice Similarity Coefficient (DSC).}
    \label{fig:ablagraph}
    \end{minipage}
    \hfill
    \begin{minipage}{0.45\textwidth}
    \centering
    \includegraphics[width=\textwidth]{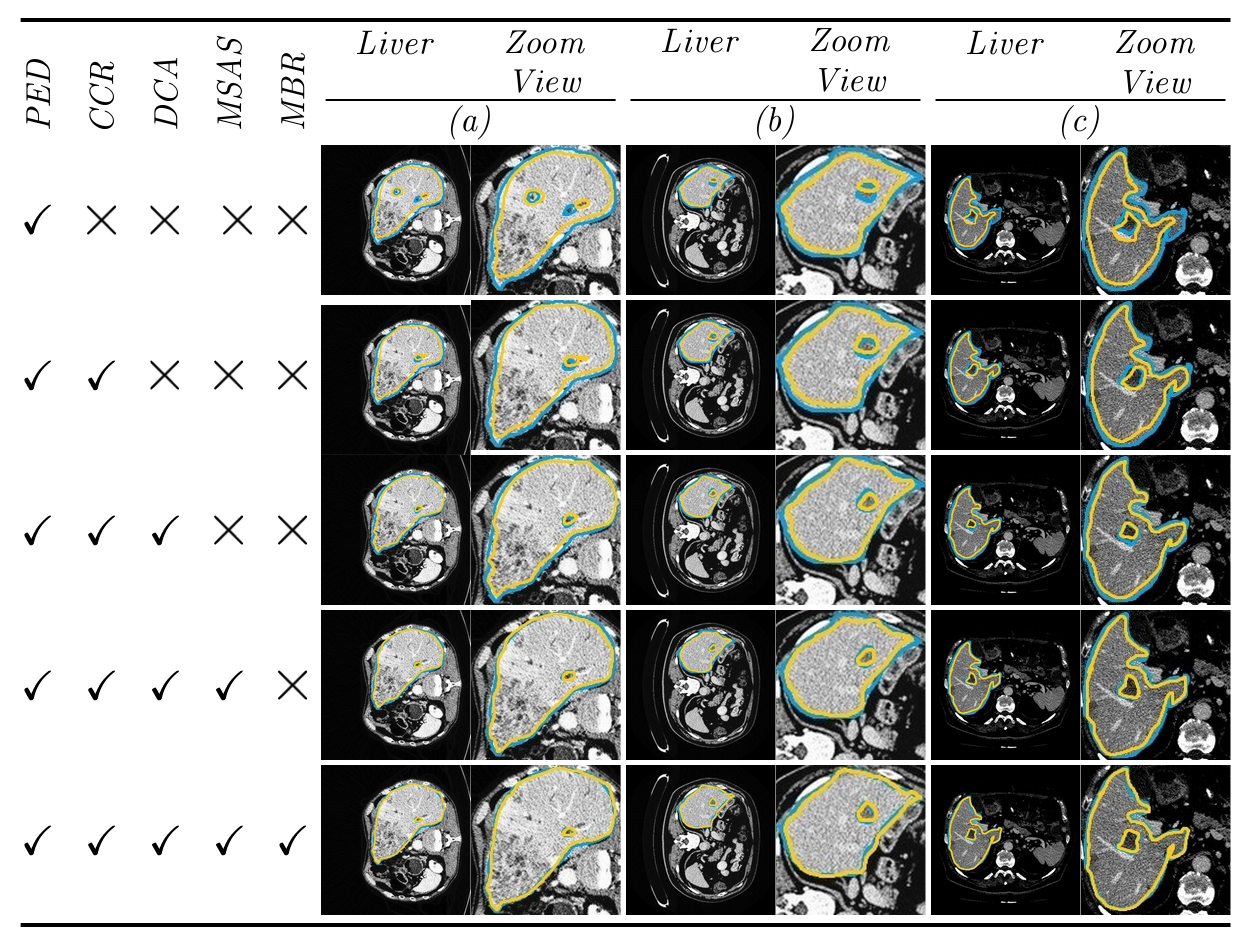}
    \caption {Visual results highlighting the contributions of essential components within the T-MPEDNet segmentation framework. (Blue boundaries - ground truth segmentation and yellow boundaries - predicted segmentation)}
    \label{fig:ablavis}
    \end{minipage}    
\end{figure}

This ablation study examines the influence of individual modules on liver and tumor segmentation. Each module contributes positively to liver segmentation, with MBR providing a minor yet positive contribution to liver DSC of 0.2\% and 0.5\% to tumor DSC. The role of MBR is to obtain accurate boundary labels for liver and tumor. Next, MSAS demonstrates its effectiveness in capturing multi-scale contextual information. Its removal caused Liver DSC to dip by 1.7\% and Tumor DSC by 2.7\%. The role of DCA is to learn the global features among local fine-grained spatial feature maps. Removing the DCA module has adverse effects on DSC; it declines the liver DSC by 2.2\% and tumor DSC by 5.2\%. CCR module leverages channel-wise dependencies in feature maps by automatically acquiring weights and recalibrating original feature channels, enhancing feature representation. Removal of the CCR module significantly drops the DSC of both liver and tumor by 4.1\% and 7.7\%, highlighting its crucial role. PED is the backbone block integrated within the T-MPEDNet framework, and it is responsible for learning adaptive recalibrates fused features leveraging channel-wise to learn fine-grained details while preserving spatial integrity. \autoref{fig:ablagraph} depicts the effects of integrating various modules within the T-MPEDNet framework on the Dice Similarity Coefficient (DSC) metric. The x-axis likely represents the sequential addition of these modules, while the y-axis represents the corresponding DSC values.

\autoref{fig:ablavis} visualizes the segmentation results on three CT scans using various configurations of the T-MPEDNet framework. The first row depicts the T-MPEDNet without additional modules. The significant deviations between the predicted segmentations (yellow boundary) and the ground truth (blue boundary) indicate that the framework inability to capture discriminative features for accurate segmentation. As more modules are incorporated in subsequent rows, the predicted segmentations align more closely with the ground truth. This progressive improvement suggests that each module contributes to enhanced feature extraction capability. The integration of these learned features ultimately leads to superior segmentation performance.

\section{Discussion}
\label{sec:discussion}
Segmentation of the liver and tumors in medical images, often obtained using CT scans or MRIs, is a critical first step in diagnosing and treating liver cancer. This process allows healthcare professionals to identify and analyze the affected areas precisely, guiding treatment decisions and improving patient outcomes. In this paper, we present an \textbf{T}ransformer-aware \textbf{M}ultiscale \textbf{P}rogressive \textbf{E}ncoder-\textbf{D}ecoder \textbf{Net}work (T-MPEDNet), aims to perform automated liver and tumor segmentation in CT scans, bridging the gap between research and clinical applications.

Recent advancements in Convolutional Neural Network (CNN) architectures and training strategies have contributed significantly to the success of convolutional-based methods in liver tumor segmentation~\citep{Guo2024eltsnet}. The convolutional operations in CNNs mainly focus on spatial dependencies and have limitations in mapping long-range dependencies. This study introduces a transformer-based dynamic contextual attention mechanism followed by a multi-scale feature extractor into progressive encoder-decoder architecture to mitigate this issue. Dynamic contextual attention helps learn the global dependencies within the adaptive features. The dynamic convolutions integrated within the attention mechanism can focus on diverse positions, enhancing its representational flexibility. The multi-scale feature extractor refines the obtained contextual features by varying the receptive field size and efficiently captures fine and coarse-grained features.
Meanwhile, considering that most existing liver and tumor segmentation methods utilize skip connections~\citep{Wang2023contextfusion} to facilitate the integration of low-level and high-level features. Our method leverages skip connection in progressive encoder architecture to recalibrate the features during upsampling by combining features from different resolutions. Additionally, our method includes the morphological erosion operation to refine the boundary further to obtain accurate boundary labels.

Experimental evaluations on the LiTS and 3DIRCADb dataset in \autoref{table:sotalits3d} indicate that the proposed T-MPEDNet is superior to twelve state-of-the-art methods and a highly competitive segmentation framework. \autoref{table:ablation} assess the individual contributions of the key modules within the T-MPEDNet framework. The Dice Similarity Coefficient (DSC) improved by 1.9\% when channel-wise feature recalibration is integrated within the progressive encoder-decoder architecture. Furthermore, the DSC is improved by  2\% when transformer-based dynamic contextual attention and multi-scale feature extractor are integrated, which signifies the essential contribution of individual modules within the T-MPEDNet framework.

While our framework demonstrates promising results, we acknowledge some limitations that warrant further exploration. Notably, the high memory consumption and computational cost of 3D networks constrained us to train and test on 2D images, neglecting the valuable inter-slice features. To mitigate the high memory requirements and computational burden associated with 3D networks, we opted for training and testing on 2D CT images. This approach, while computationally efficient, does not capture the inherent 3D context present in CT scans, potentially limiting the model's ability to leverage inter-slice information. Additionally, down-sampling the input size to $256 \times 256$ may lead to the loss of crucial spatial information from CT scans. Future work will explore the development of a more flexible framework that effectively extracts 3D context for both liver and tumor structures. This refined approach holds the potential to significantly enhance the accuracy and efficiency of automated liver and tumor segmentation.

\section{Conclusion}
\label{sec:conclusion}
In this study, we present a novel \textbf{T}ransformer-aware \textbf{M}ultiscale \textbf{P}rogressive \textbf{E}ncoder-\textbf{D}ecoder \textbf{Net}work (T-MPEDNet) for automated liver and tumor segmentation from CT scans. T-MPEDNet harnesses the power of a Transformer-inspired dynamic attention mechanism coupled with a multi-scale feature extractor. This synergistic approach enables the capture of fine-grained features utilizing a larger receptive field while preserving global features. The dynamic attention mechanism learns the contextual dependencies within the hepatic tumor and non-tumor regions by dynamically adjusting the convolution kernel, enabling the mapping of spatial relations. Subsequently, the multi-scale feature extractor enhances the global feature map's contextual understanding of liver and tumor anatomy. Additionally, morphological operations are incorporated to refine liver and tumor boundaries, ensuring precise and accurate segmentation. Extensive evaluations on LiTS and 3DIRCADb datasets demonstrate the superiority of T-MPEDNet compared to twelve state-of-the-art methods. The proposed framework effectively automates the segmentation of intricate and small liver and tumor boundaries in CT scans, presenting valuable applications in clinical procedures such as treatment planning for hepatic diseases and tumors.


\bibliographystyle{elsarticle-num} 
\bibliography{refs}

\begin{thebibliography}{10}
\expandafter\ifx\csname url\endcsname\relax
  \def\url#1{\texttt{#1}}\fi
\expandafter\ifx\csname urlprefix\endcsname\relax\def\urlprefix{URL }\fi
\expandafter\ifx\csname href\endcsname\relax
  \def\href#1#2{#2} \def\path#1{#1}\fi

\bibitem{Ma2021abdomenCT}
J.~Ma, Y.~Zhang, S.~Gu, C.~Zhu, C.~Ge, Y.~Zhang, X.~An, C.~Wang, Q.~Wang, X.~Liu, S.~Cao, Q.~Zhang, S.~Liu, Y.~Wang, Y.~Li, J.~He, X.~Yang, Abdomenct-1k: Is abdominal organ segmentation a solved problem, IEEE Transactions on Pattern Analysis and Machine Intelligence (2021).
\newblock \href {https://doi.org/10.1109/TPAMI.2021.3100536} {\path{doi:10.1109/TPAMI.2021.3100536}}.

\bibitem{Valanarasu2022kiunet}
J.~M.~J. Valanarasu, V.~A. Sindagi, I.~Hacihaliloglu, V.~M. Patel, Kiu-net: Overcomplete convolutional architectures for biomedical image and volumetric segmentation, IEEE Transactions on Medical Imaging 41 (2022).
\newblock \href {https://doi.org/10.1109/TMI.2021.3130469} {\path{doi:10.1109/TMI.2021.3130469}}.

\bibitem{Li2018hdenseunet}
X.~Li, H.~Chen, X.~Qi, Q.~Dou, C.~W. Fu, P.~A. Heng, H-denseunet: Hybrid densely connected unet for liver and tumor segmentation from ct volumes, IEEE Transactions on Medical Imaging 37 (2018).
\newblock \href {https://doi.org/10.1109/TMI.2018.2845918} {\path{doi:10.1109/TMI.2018.2845918}}.

\bibitem{Yang2021region}
Z.~Yang, Y.~qian Zhao, M.~Liao, S.~hu~Di, Y.~zhan Zeng, Semi-automatic liver tumor segmentation with adaptive region growing and graph cuts, Biomedical Signal Processing and Control 68 (2021).
\newblock \href {https://doi.org/10.1016/j.bspc.2021.102670} {\path{doi:10.1016/j.bspc.2021.102670}}.

\bibitem{He2024deform}
W.~He, C.~Zhang, J.~Dai, L.~Liu, T.~Wang, X.~Liu, Y.~Jiang, N.~Li, J.~Xiong, L.~Wang, Y.~Xie, X.~Liang, A statistical deformation model-based data augmentation method for volumetric medical image segmentation, Medical Image Analysis 91 (2024) 102984.
\newblock \href {https://doi.org/10.1016/j.media.2023.102984} {\path{doi:10.1016/j.media.2023.102984}}.

\bibitem{Liu2022threshold}
W.~Liu, H.~Yang, T.~Tian, Z.~Cao, X.~Pan, W.~Xu, Y.~Jin, F.~Gao, Full-resolution network and dual-threshold iteration for retinal vessel and coronary angiograph segmentation, IEEE Journal of Biomedical and Health Informatics 26 (2022).
\newblock \href {https://doi.org/10.1109/JBHI.2022.3188710} {\path{doi:10.1109/JBHI.2022.3188710}}.

\bibitem{Dickson2024sparse}
A.~J. Dickson, J.~A. Linsely, V.~A.~A. Daniel, K.~Rahul, Sparse deep belief network coupled with extended local fuzzy active contour model-based liver cancer segmentation from abdomen ct images, Medical \& Biological Engineering \& Computing (2024) 1--14\href {https://doi.org/10.1007/s11517-023-03001-y} {\path{doi:10.1007/s11517-023-03001-y}}.

\bibitem{Di2023tdnet}
S.~Di, Y.~Q. Zhao, M.~Liao, F.~Zhang, X.~Li, Td-net: A hybrid end-to-end network for automatic liver tumor segmentation from ct images, IEEE Journal of Biomedical and Health Informatics 27 (2023).
\newblock \href {https://doi.org/10.1109/JBHI.2022.3181974} {\path{doi:10.1109/JBHI.2022.3181974}}.

\bibitem{Ronneberger2015unet}
O.~Ronneberger, P.~Fischer, T.~Brox, U-net: Convolutional networks for biomedical image segmentation, in: Lecture Notes in Computer Science (including subseries Lecture Notes in Artificial Intelligence and Lecture Notes in Bioinformatics), Vol. 9351, 2015, pp. 234--241.
\newblock \href {https://doi.org/10.1007/978-3-319-24574-4_28} {\path{doi:10.1007/978-3-319-24574-4_28}}.

\bibitem{Liao2024mscfunet}
M.~Liao, H.~Tang, X.~Li, P.~Vijayakumar, V.~Arya, B.~B. Gupta, A lightweight network for abdominal multi-organ segmentation based on multi-scale context fusion and dual self-attention, Information Fusion 108 (2024) 102401.
\newblock \href {https://doi.org/10.1016/j.inffus.2024.102401} {\path{doi:10.1016/j.inffus.2024.102401}}.

\bibitem{Tu2023slice}
D.~Y. Tu, P.~C. Lin, H.~H. Chou, M.~R. Shen, S.~Y. Hsieh, Slice-fusion: Reducing false positives in liver tumor detection for mask r-cnn, IEEE/ACM Transactions on Computational Biology and Bioinformatics 20 (2023).
\newblock \href {https://doi.org/10.1109/TCBB.2023.3265394} {\path{doi:10.1109/TCBB.2023.3265394}}.

\bibitem{Li2023ace}
Y.~Li, B.~Zou, P.~Dai, M.~Liao, H.~X. Bai, Z.~Jiao, Ac-e network: Attentive context-enhanced network for liver segmentation, IEEE Journal of Biomedical and Health Informatics 27 (2023).
\newblock \href {https://doi.org/10.1109/JBHI.2023.3278079} {\path{doi:10.1109/JBHI.2023.3278079}}.

\bibitem{Kushnure2022hfru}
D.~T. Kushnure, S.~N. Talbar, Hfru-net: High-level feature fusion and recalibration unet for automatic liver and tumor segmentation in ct images, Computer Methods and Programs in Biomedicine 213 (2022).
\newblock \href {https://doi.org/10.1016/j.cmpb.2021.106501} {\path{doi:10.1016/j.cmpb.2021.106501}}.

\bibitem{Jiang2023rmau}
L.~Jiang, J.~Ou, R.~Liu, Y.~Zou, T.~Xie, H.~Xiao, T.~Bai, Rmau-net: Residual multi-scale attention u-net for liver and tumor segmentation in ct images, Computers in Biology and Medicine 158 (2023).
\newblock \href {https://doi.org/10.1016/j.compbiomed.2023.106838} {\path{doi:10.1016/j.compbiomed.2023.106838}}.

\bibitem{Liu2022transunetp}
Y.~Liu, H.~Wang, Z.~Chen, K.~Huangliang, H.~Zhang, Transunet+: Redesigning the skip connection to enhance features in medical image segmentation, Knowledge-Based Systems 256 (2022).
\newblock \href {https://doi.org/10.1016/j.knosys.2022.109859} {\path{doi:10.1016/j.knosys.2022.109859}}.

\bibitem{Xie2021cotr}
Y.~Xie, J.~Zhang, C.~Shen, Y.~Xia, Cotr: Efficiently bridging cnn and transformer for 3d medical image segmentation, in: Medical Image Computing and Computer Assisted Intervention--MICCAI 2021: 24th International Conference, Strasbourg, France, September 27--October 1, 2021, Proceedings, Part III 24, Vol. 12903 LNCS, 2021, pp. 171--180.
\newblock \href {https://doi.org/10.1007/978-3-030-87199-4_16} {\path{doi:10.1007/978-3-030-87199-4_16}}.

\bibitem{Vaswani2017}
A.~Vaswani, N.~Shazeer, N.~Parmar, J.~Uszkoreit, L.~Jones, A.~N. Gomez, {\L}.~Kaiser, I.~Polosukhin, Attention is all you need, Advances in neural information processing systems 30 (2017).

\bibitem{Zhang2024fafsunet}
X.~Zhang, S.~Yang, Y.~Jiang, Y.~Chen, F.~Sun, Fafs-unet: Redesigning skip connections in unet with feature aggregation and feature selection, Computers in Biology and Medicine 170 (2024) 108009.
\newblock \href {https://doi.org/10.1016/j.compbiomed.2024.108009} {\path{doi:10.1016/j.compbiomed.2024.108009}}.

\bibitem{Bilic2023lits}
P.~Bilic, P.~Christ, H.~B. Li, E.~Vorontsov, A.~Ben-Cohen, G.~Kaissis, A.~Szeskin, C.~Jacobs, G.~E.~H. Mamani, G.~Chartrand, F.~Lohöfer, J.~W. Holch, W.~Sommer, F.~Hofmann, A.~Hostettler, N.~Lev-Cohain, M.~Drozdzal, M.~M. Amitai, R.~Vivanti, J.~Sosna, I.~Ezhov, A.~Sekuboyina, F.~Navarro, F.~Kofler, J.~C. Paetzold, S.~Shit, X.~Hu, J.~Lipková, M.~Rempfler, M.~Piraud, J.~Kirschke, B.~Wiestler, Z.~Zhang, C.~Hülsemeyer, M.~Beetz, F.~Ettlinger, M.~Antonelli, W.~Bae, M.~Bellver, L.~Bi, H.~Chen, G.~Chlebus, E.~B. Dam, Q.~Dou, C.~W. Fu, B.~Georgescu, X.~G. i~Nieto, F.~Gruen, X.~Han, P.~A. Heng, J.~Hesser, J.~H. Moltz, C.~Igel, F.~Isensee, P.~Jäger, F.~Jia, K.~C. Kaluva, M.~Khened, I.~Kim, J.~H. Kim, S.~Kim, S.~Kohl, T.~Konopczynski, A.~Kori, G.~Krishnamurthi, F.~Li, H.~Li, J.~Li, X.~Li, J.~Lowengrub, J.~Ma, K.~Maier-Hein, K.~K. Maninis, H.~Meine, D.~Merhof, A.~Pai, M.~Perslev, J.~Petersen, J.~Pont-Tuset, J.~Qi, X.~Qi, O.~Rippel, K.~Roth, I.~Sarasua, A.~Schenk, Z.~Shen, J.~Torres, C.~Wachinger, C.~Wang,
  L.~Weninger, J.~Wu, D.~Xu, X.~Yang, S.~C.~H. Yu, Y.~Yuan, M.~Yue, L.~Zhang, J.~Cardoso, S.~Bakas, R.~Braren, V.~Heinemann, C.~Pal, A.~Tang, S.~Kadoury, L.~Soler, B.~van Ginneken, H.~Greenspan, L.~Joskowicz, B.~Menze, The liver tumor segmentation benchmark (lits), Medical Image Analysis 84 (2023) 102680.
\newblock \href {https://doi.org/10.1016/J.MEDIA.2022.102680} {\path{doi:10.1016/J.MEDIA.2022.102680}}.

\bibitem{soler20103d}
L.~Soler, A.~Hostettler, V.~Agnus, A.~Charnoz, J.-B. Fasquel, J.~Moreau, A.-B. Osswald, M.~Bouhadjar, J.~Marescaux, 3d image reconstruction for comparison of algorithm database, URL: https://www. ircad. fr/research/data-sets/liver-segmentation-3d-ircadb-01 (2010).

\bibitem{Gopinath2023}
B.~Gopinath, V.~V. Kumar, H.~Kag, K.~Joshi, A.~Rajput, D.~Kapila, Medical image segmentation using deep learning techniques: Advancement and challenges, in: 2023 3rd International Conference on Advance Computing and Innovative Technologies in Engineering, ICACITE 2023, 2023, pp. 1088--1093.
\newblock \href {https://doi.org/10.1109/ICACITE57410.2023.10183260} {\path{doi:10.1109/ICACITE57410.2023.10183260}}.

\bibitem{Mahendran2023}
N.~Mahendran, P.~Muthuvel, T.~Arunprasath, M.~P. Rajasekaran, J.~B. Nirmala, R.~Kottaimalai, Identification and segmentation of tumour in brain mri using deep learning techniques, in: Proceedings of the 2023 2nd International Conference on Electronics and Renewable Systems, ICEARS 2023, 2023, pp. 1214--1219.
\newblock \href {https://doi.org/10.1109/ICEARS56392.2023.10085383} {\path{doi:10.1109/ICEARS56392.2023.10085383}}.

\bibitem{Soomro2023}
T.~A. Soomro, L.~Zheng, A.~J. Afifi, A.~Ali, S.~Soomro, M.~Yin, J.~Gao, Image segmentation for mr brain tumor detection using machine learning: A review (2023).
\newblock \href {https://doi.org/10.1109/RBME.2022.3185292} {\path{doi:10.1109/RBME.2022.3185292}}.

\bibitem{Qayyum2024coattunet}
A.~Qayyum, I.~Razzak, M.~Mazher, X.~Lu, S.~A. Niederer, Unsupervised unpaired multiple fusion adaptation aided with self-attention generative adversarial network for scar tissues segmentation framework, Information Fusion 106 (2024).
\newblock \href {https://doi.org/10.1016/j.inffus.2024.102226} {\path{doi:10.1016/j.inffus.2024.102226}}.

\bibitem{Di2022autolt}
S.~Di, Y.~Zhao, M.~Liao, Z.~Yang, Y.~Zeng, Automatic liver tumor segmentation from ct images using hierarchical iterative superpixels and local statistical features, Expert Systems with Applications 203 (2022).
\newblock \href {https://doi.org/10.1016/j.eswa.2022.117347} {\path{doi:10.1016/j.eswa.2022.117347}}.

\bibitem{Chen2023draunet}
Y.~Chen, C.~Zheng, T.~Zhou, L.~Feng, L.~Liu, Q.~Zeng, G.~Wang, A deep residual attention-based u-net with a biplane joint method for liver segmentation from ct scans, Computers in Biology and Medicine 152 (2023).
\newblock \href {https://doi.org/10.1016/j.compbiomed.2022.106421} {\path{doi:10.1016/j.compbiomed.2022.106421}}.

\bibitem{ChenKiu2023}
G.~Chen, Z.~Li, J.~Wang, J.~Wang, S.~Du, J.~Zhou, J.~Shi, Y.~Zhou, An improved 3d kiu-net for segmentation of liver tumor, Computers in Biology and Medicine 160 (2023).
\newblock \href {https://doi.org/10.1016/j.compbiomed.2023.107006} {\path{doi:10.1016/j.compbiomed.2023.107006}}.

\bibitem{LI2020}
S.~LI, G.~K. TSO, K.~HE, Bottleneck feature supervised u-net for pixel-wise liver and tumor segmentation, Expert Systems with Applications 145 (2020).
\newblock \href {https://doi.org/10.1016/j.eswa.2019.113131} {\path{doi:10.1016/j.eswa.2019.113131}}.

\bibitem{Zhao2022}
J.~Zhao, X.~Hou, M.~Pan, H.~Zhang, Attention-based generative adversarial network in medical imaging: A narrative review, Computers in Biology and Medicine 149 (2022).
\newblock \href {https://doi.org/10.1016/j.compbiomed.2022.105948} {\path{doi:10.1016/j.compbiomed.2022.105948}}.

\bibitem{Li2023sdmt}
X.~Li, S.~Lv, M.~Li, J.~Zhang, Y.~Jiang, Y.~Qin, H.~Luo, S.~Yin, Sdmt: Spatial dependence multi-task transformer network for 3d knee mri segmentation and landmark localization, IEEE Transactions on Medical Imaging 42 (2023).
\newblock \href {https://doi.org/10.1109/TMI.2023.3247543} {\path{doi:10.1109/TMI.2023.3247543}}.

\bibitem{Tomar2023}
N.~K. Tomar, D.~Jha, M.~A. Riegler, H.~D. Johansen, D.~Johansen, J.~Rittscher, P.~Halvorsen, S.~Ali, Fanet: A feedback attention network for improved biomedical image segmentation, IEEE Transactions on Neural Networks and Learning Systems 34 (2023).
\newblock \href {https://doi.org/10.1109/TNNLS.2022.3159394} {\path{doi:10.1109/TNNLS.2022.3159394}}.

\bibitem{Wu2023}
Y.~Wu, Q.~Kong, L.~Zhang, A.~Castiglione, M.~Nappi, S.~Wan, Cdt-cad: Context-aware deformable transformers for end-to-end chest abnormality detection on x-ray images, IEEE/ACM Transactions on Computational Biology and Bioinformatics (2023).
\newblock \href {https://doi.org/10.1109/TCBB.2023.3258455} {\path{doi:10.1109/TCBB.2023.3258455}}.

\bibitem{Li2022litsnet}
J.~Li, G.~Huang, J.~He, Z.~Chen, C.~M. Pun, Z.~Yu, W.~K. Ling, L.~Liu, J.~Zhou, J.~Huang, Shift-channel attention and weighted-region loss function for liver and dense tumor segmentation, Medical Physics 49 (2022).
\newblock \href {https://doi.org/10.1002/mp.15816} {\path{doi:10.1002/mp.15816}}.

\bibitem{Liu2023}
H.~Liu, Y.~Fu, S.~Zhang, J.~Liu, Y.~Wang, G.~Wang, J.~Fang, Gcha-net: Global context and hybrid attention network for automatic liver segmentation, Computers in Biology and Medicine 152 (2023).
\newblock \href {https://doi.org/10.1016/j.compbiomed.2022.106352} {\path{doi:10.1016/j.compbiomed.2022.106352}}.

\bibitem{Alam2023}
M.~S. Alam, D.~Wang, Q.~Liao, A.~Sowmya, A multi-scale context aware attention model for medical image segmentation, IEEE Journal of Biomedical and Health Informatics 27 (2023).
\newblock \href {https://doi.org/10.1109/JBHI.2022.3227540} {\path{doi:10.1109/JBHI.2022.3227540}}.

\bibitem{Cui2021}
B.~Cui, W.~Jing, L.~Huang, Z.~Li, Y.~Lu, Sanet: A sea-land segmentation network via adaptive multiscale feature learning, IEEE Journal of Selected Topics in Applied Earth Observations and Remote Sensing 14 (2021).
\newblock \href {https://doi.org/10.1109/JSTARS.2020.3040176} {\path{doi:10.1109/JSTARS.2020.3040176}}.

\bibitem{Du2021}
X.~Du, J.~Wang, W.~Sun, Densely connected u-net retinal vessel segmentation algorithm based on multi-scale feature convolution extraction, Medical Physics 48 (2021).
\newblock \href {https://doi.org/10.1002/mp.14944} {\path{doi:10.1002/mp.14944}}.

\bibitem{Abdar2023}
M.~Abdar, M.~A. Fahami, L.~Rundo, P.~Radeva, A.~F. Frangi, U.~R. Acharya, A.~Khosravi, H.~K. Lam, A.~Jung, S.~Nahavandi, Hercules: Deep hierarchical attentive multilevel fusion model with uncertainty quantification for medical image classification, IEEE Transactions on Industrial Informatics 19 (2023).
\newblock \href {https://doi.org/10.1109/TII.2022.3168887} {\path{doi:10.1109/TII.2022.3168887}}.

\bibitem{Yue2023}
G.~Yue, S.~Li, R.~Cong, T.~Zhou, B.~Lei, T.~Wang, Attention-guided pyramid context network for polyp segmentation in colonoscopy images, IEEE Transactions on Instrumentation and Measurement 72 (2023).
\newblock \href {https://doi.org/10.1109/TIM.2023.3244219} {\path{doi:10.1109/TIM.2023.3244219}}.

\bibitem{Dar2025}
S.~S. Dar, B.~Kaurav, A.~Jain, C.~S. Raghaw, M.~Z.~U. Rehman, N.~Kumar, An explainable deep neural network with frequency-aware channel and spatial refinement for flood prediction in sustainable cities, Sustainable Cities and Society 130 (2025) 106480.
\newblock \href {https://doi.org/10.1016/j.scs.2025.106480} {\path{doi:10.1016/j.scs.2025.106480}}.

\bibitem{Lei2022defednet}
T.~Lei, R.~Wang, Y.~Zhang, Y.~Wan, C.~Liu, A.~K. Nandi, Defed-net: Deformable encoder-decoder network for liver and liver tumor segmentation, IEEE Transactions on Radiation and Plasma Medical Sciences 6 (2022).
\newblock \href {https://doi.org/10.1109/TRPMS.2021.3059780} {\path{doi:10.1109/TRPMS.2021.3059780}}.

\bibitem{Chen2022fraunet}
Y.~Chen, C.~Zheng, F.~Hu, T.~Zhou, L.~Feng, G.~Xu, Z.~Yi, X.~Zhang, Efficient two-step liver and tumour segmentation on abdominal ct via deep learning and a conditional random field, Computers in Biology and Medicine 150 (2022).
\newblock \href {https://doi.org/10.1016/j.compbiomed.2022.106076} {\path{doi:10.1016/j.compbiomed.2022.106076}}.

\bibitem{Chen2023msfanet}
Y.~Chen, C.~Zheng, W.~Zhang, H.~Lin, W.~Chen, G.~Zhang, G.~Xu, F.~Wu, Ms-fanet: Multi-scale feature attention network for liver tumor segmentation, Computers in Biology and Medicine 163 (2023).
\newblock \href {https://doi.org/10.1016/j.compbiomed.2023.107208} {\path{doi:10.1016/j.compbiomed.2023.107208}}.

\bibitem{Lv2022}
P.~Lv, J.~Wang, X.~Zhang, C.~Shi, Deep supervision and atrous inception-based u-net combining crf for automatic liver segmentation from ct, Scientific Reports 12 (2022).
\newblock \href {https://doi.org/10.1038/s41598-022-21562-0} {\path{doi:10.1038/s41598-022-21562-0}}.

\bibitem{Guo2024eltsnet}
X.~Guo, Z.~Wang, P.~Wu, Y.~Li, F.~E. Alsaadi, N.~Zeng, Elts-net: An enhanced liver tumor segmentation network with augmented receptive field and global contextual information, Computers in Biology and Medicine 169 (2024) 107879.
\newblock \href {https://doi.org/10.1016/j.compbiomed.2023.107879} {\path{doi:10.1016/j.compbiomed.2023.107879}}.

\bibitem{Wang2023contextfusion}
Z.~Wang, J.~Zhu, S.~Fu, Y.~Ye, Context fusion network with multi-scale-aware skip connection and twin-split attention for liver tumor segmentation, Medical and Biological Engineering and Computing 61 (2023).
\newblock \href {https://doi.org/10.1007/s11517-023-02876-1} {\path{doi:10.1007/s11517-023-02876-1}}.

\end{thebibliography}

\end{document}